\documentclass[10pt]{article} 
\usepackage[accepted]{tmlr}

\usepackage{amsmath,amsfonts,bm}

\def\eqref#1{equation~\ref{#1}}

\def\1{\bm{1}}

\DeclareMathAlphabet{\mathsfit}{\encodingdefault}{\sfdefault}{m}{sl}
\SetMathAlphabet{\mathsfit}{bold}{\encodingdefault}{\sfdefault}{bx}{n}

\usepackage[utf8]{inputenc} 
\usepackage[T1]{fontenc}    
\usepackage{url}            
\usepackage{booktabs}       
\usepackage{amsfonts}       
\usepackage{nicefrac}       
\usepackage{microtype}      
\usepackage{xcolor}         
\usepackage{tabularx}
\usepackage{subcaption}
\usepackage{xspace}
\usepackage{multicol}
\usepackage{multirow}
\usepackage{tabularray}
\usepackage{amsmath}
\usepackage{algorithm}%
\usepackage{algorithmicx}%
\usepackage{algpseudocode}
\usepackage{amsfonts}
\usepackage{siunitx}
\usepackage{colortbl}
\usepackage{graphicx}
\usepackage{lipsum}
\usepackage{nicefrac}
\usepackage{tabularray}
\usepackage{wrapfig}

\usepackage{pgfplots,gincltex}
\usepgfplotslibrary{groupplots}
\pgfplotsset{compat=1.6}
\usepackage{tikz}
\usetikzlibrary{arrows.meta}

\usepackage[pagebackref,breaklinks,colorlinks,allcolors=blue]{hyperref}
\usepackage{url}

\newcommand{\ourmethod}{AnyIR }
\definecolor{pltblue}{RGB}{174, 199, 232}
\definecolor{pltorange}{RGB}{255, 229, 204}
\definecolor{pltgreen}{RGB}{204, 229, 204}
\definecolor{pltred}{RGB}{229, 204, 204}
\definecolor{pltpurple}{RGB}{239, 218, 230}

\definecolor{tabblue}{HTML}{1f77b4}
\definecolor{taborange}{HTML}{ff7f0e}
\definecolor{tabgreen}{HTML}{2ca02c}
\definecolor{tabred}{HTML}{d62728}
\definecolor{tabpurple}{HTML}{9467bd}

\definecolor{cblue}{RGB}{173, 201, 233}
\definecolor{clblue}{RGB}{222, 234, 246}
\definecolor{corange}{RGB}{255, 152, 67}
\definecolor{lorgange}{RGB}{255, 221, 149}

\def\eg{\emph{e.g.}\xspace} 
\def\ie{\emph{i.e.}\xspace} 
\def\etc{\emph{etc.}\xspace}

\newcommand{\cc}[1]{\cellcolor{clblue!50}{#1}}

\newcommand\sotaa{\textcolor{tabred}}

\title{Any Image Restoration via Efficient Spatial-Frequency Degradation Adaptation}

\author{Bin Ren\textsuperscript{1,2}\thanks{~This work was partially conducted during the visiting stay at INSAIT.} \quad
Eduard Zamfir\textsuperscript{3} \quad
Zongwei Wu\textsuperscript{3}\footnotemark[2]\thanks{~Corresponding author.} \quad
Yawei Li\textsuperscript{4} \quad 
Yidi Li\textsuperscript{5} \quad
Danda Pani Paudel\textsuperscript{6}\\
Radu Timofte\textsuperscript{3} \quad
Ming-Hsuan Yang\textsuperscript{7} \quad
Luc Van Gool\textsuperscript{6} \quad 
Nicu Sebe\textsuperscript{1} \quad 
\\[1ex]
\addr \textsuperscript{1}University of Trento, IT \quad
\textsuperscript{2}Mohamed bin Zayed University of Artificial Intelligence, UAE\\
\textsuperscript{3}University of W\"urzburg, DE \quad
\textsuperscript{4}ETH Z\"urich, CH \quad 
\textsuperscript{5}Taiyuan University of Technology, CN \\
\textsuperscript{6}INSAIT, Sofia University ``St. Kliment Ohridski'', BG \quad
\textsuperscript{7}University of California, Merced, USA
}

\def\month{02}  
\def\year{2026} 
\def\openreview{\url{https://openreview.net/forum?id=RObgCpdDqr
}} 

\begin{document}
\maketitle

\begin{abstract}
Restoring multiple degradations efficiently via just one model has become increasingly significant and impactful, especially with the proliferation of mobile devices. Traditional solutions typically involve training dedicated models per degradation, resulting in inefficiency and redundancy. More recent approaches either introduce additional modules to learn visual prompts, significantly increasing the size of the model, or incorporate cross-modal transfer from large language models trained on vast datasets, adding complexity to the system architecture. 
In contrast, our approach, termed AnyIR, takes a unified path that leverages inherent similarity across various degradations to enable both efficient and comprehensive restoration through a joint embedding mechanism, without scaling up the model or relying on large language models.
Specifically, we examine the sub-latent space of each input, identifying key components and reweighting them first in a gated manner. 
To unify intrinsic degradation awareness with contextualized attention, we propose a spatial–frequency parallel fusion strategy that strengthens spatially informed local–global interactions and enriches restoration fidelity from the frequency domain. Comprehensive evaluations across four all-in-one restoration benchmarks demonstrate that AnyIR attains state-of-the-art performance while reducing model parameters by 84\% and FLOPs by 80\% relative to the baseline. These results highlight the potential of AnyIR as an effective and lightweight solution for further all-in-one image restoration. Our code is available at: \href{https://github.com/Amazingren/AnyIR}{\textcolor{cyan}{\texttt{https://github.com/Amazingren/AnyIR}}}.
\end{abstract}
\section{Introduction}
\label{sec:introduction}
Image restoration (\ie, IR) aims to recover a clean image from its degraded observation, and a central challenge is how to handle multiple and diverse degradations within a unified framework.
This problem is inherently ill-posed since multiple solutions may explain the same output, and thus effective priors or learned representations are crucial for successful restoration. 
In real-world scenarios, degradations arise from heterogeneous sources (noise, blur, compression, weather artifacts, \etc), and practical systems must cope with them within a single pipeline rather than as isolated tasks.
Meanwhile, multi-step or per-degradation pipelines introduce storage, routing, and computation overhead during both training and deployment, whereas a single-checkpoint model better matches mobile and edge constraints and simplifies system integration.
This motivates the pursuit of \textit{a single efficient model capable of addressing multiple degradation types}.

Despite significant advances, existing IR methods still struggle to efficiently handle diverse degradations while preserving essential details~\cite{li2023efficient,ren2024ninth,ren2024sharing,wu2024compacter,yin2024flexir,zheng2024deep,ding2024masked,wu2024diffusion,li2025fractal}. 
{Many current solutions rely on per-degradation models or multi-stage pipelines, which introduce storage, routing, and computation overhead during both training and deployment, and are difficult to scale or deploy on mobile and edge platforms.}
In contrast, a single-checkpoint model better matches deployment constraints and simplifies system integration. 
{This work addresses this gap by pursuing an efficient All-in-one IR model that can handle multiple degradation types within a unified representation-learning framework.}

The complexity of restoration models~\cite{liang2021swinir,zamir2022restormer,zamir2021multi,wang2022uformer,chen2022simple} arises primarily from the diversity of degradations. Consequently, most methods are tailored to specific tasks with limited generalization. A versatile system often requires integrating multiple specialized models, leading to heavy frameworks. Some studies have shown that a single architecture can handle multiple degradations, but they lack parameter unification, producing multiple checkpoints for different tasks~\cite{chen2022simple,ren2024sharing}. This reduces efficiency despite simplifying the system. 
Recent research has sought architectural and parameter unification~\cite{li2022all,potlapalli2023promptir,zhang2023ingredient,liu2022tape}. Diffusion-based methods demonstrate strong generative capacity~\cite{ren2023multiscale,jiang2023autodir,zhao2024denoising}, while prompt-based designs guide networks with modality-specific embeddings~\cite{wang2023promptrestorer,potlapalli2023promptir,li2023prompt}. 
Other works leverage text as an intermediate representation~\cite{luo2023controlling,conde2024high}. Despite promising results, these approaches significantly increase model size, inference time, or rely on degradation-specific supervision.

In this paper, we take a unified perspective: \textit{although each degradation has its characteristics, all restoration tasks share underlying principles of separating nuisance degradations from structural image information}. From a learning viewpoint, this setting can be regarded as a \emph{multi-environment learning problem}, where degradations correspond to environments 
and the objective is to learn a representation 
that is invariant to degradation-specific nuisances yet sufficient for restoration.
Unlike large-scale priors in LLMs, our method builds invariance directly from single-image cues, yielding both efficiency and strong generalization. 
To this end, we propose AnyIR, a lightweight framework that integrates global context and local degradation-aware cues. Specifically, we introduce a gated local block that disentangles fine-grained degradation-aware details (ego, shifted, and scaled parts), adaptively reweighted via gating, and a parallel attention pathway to capture global dependencies. A spatial–frequency fusion mechanism further intertwines the two representations, balancing structural integrity with fine detail recovery. Importantly, features are processed in sub-latent partitions before aggregation, reducing computational cost while retaining rich information. This design enables AnyIR to act as an implicit degradation-invariant learner, effective and efficient across diverse restoration settings (see Fig.~\ref{fig:teaser}).

Our main contributions are summarized as follows:
\begin{itemize}
    \item We propose \ourmethod, a unified and efficient all-in-one IR model that achieves superior performance while reducing computational cost by \textbf{85.6\%} compared to state-of-the-art counterparts.
    \item We design a novel local-global gated intertwining mechanism combined with a spatial–frequency fusion strategy, enabling cohesive and adaptive embeddings without degradation-specific supervision.
    \item Through extensive experiments on diverse restoration tasks, we demonstrate the effectiveness and efficiency of \ourmethod, providing a strong and practical baseline for future research in all-in-one IR.
\end{itemize}

\section{Related Work}
\label{sec:related-work}
\paragraph{Image Restoration (IR)}
Image restoration aims to solve a highly ill-posed problem by reconstructing high-quality images from their degraded counterparts. Due to its importance, IR has been applied to various applications~\cite{richardson1972bayesian,banham1997digital,li2023lsdir,zamfir2024details,miao2024scenegraphloc,zheng2024textual}. Initially, IR was addressed through model-based solutions involving the search for solutions to specific formulations. However, learning-based approaches have gained much attention with the significant advancements in deep neural networks. Numerous approaches have been developed, including regression-based~\cite{lim2017enhanced,lai2017deep,liang2021swinir,chen2021learning,li2023efficient,zhang2024transcending} and generative model-based pipelines~\cite{gao2023implicit,wang2022zero,luo2023image,yue2023resshift,zhao2024denoising,liu2023spatio} that are based on convolutional~\cite{dong2015compression,zhang2017learning,zhang2017beyond,wang2018recovering}, MLP~\cite{tu2022maxim}, state-space mode~\cite{guo2024mambair,zhu2024vision,mamba,mamba2}, or vision transformers-based (ViTs) architectures~\cite{liang2021swinir,li2023efficient,zamir2022restormer,ren2023masked,dosovitskiy2020image}. 
Although state-of-the-art methods have achieved promising performance, mainstream IR solutions still focus on addressing single degradation tasks such as denoising~\cite{zhang2017learning,zhang2019residual}, dehazing~\cite{ren2020single,wu2021contrastive}, deraining~\cite{jiang2020multi,ren2019progressive}, and deblurring~\cite{kong2023efficient,ren2023multiscale}.

\paragraph{One for Any Image Restoration}
Training a task-specific model to address a single degradation is effective, but impractical for real-world deployment due to the need for separate models for each corruption. 
In practice, images often suffer from multiple, overlapping degradations, making it inefficient to address them individually. Such task-specific solutions require significant computing and storage resources, and their environmental cost grows with the number of degradations. 
To overcome these limitations, the emerging \emph{All-in-One} IR setting develops a \emph{single blind model} that handles multiple degradations simultaneously, without requiring task-specific specialization~\cite{tang2025degradation,jiang2025survey}. 
Moreover, training a unified model across multiple degradations enables the learning of transferable structures, which facilitates adaptation to new or mixed degradations through lightweight fine-tuning rather than training separate models from scratch.
From a learning perspective, this setting can also be interpreted as a form of multi-environment learning, where different degradations act as environments and the challenge is to learn representations that are robust across them while still sufficient for reconstruction.
Several representative approaches have been proposed. AirNet~\cite{li2022all} employs contrastive learning to derive degradation-aware embeddings from corrupted inputs, which are then used to guide restoration. IDR~\cite{zhang2023ingredient} follows a meta-learning perspective, decomposing degradations into fundamental physical principles and adapting via a two-stage learning process. More recently, the prompt-based paradigm~\cite{potlapalli2023promptir,wang2023promptrestorer,li2023prompt} has been introduced, where learned visual prompts condition a single model across degradations. Such prompts act as task embeddings that steer the network, with extensions including frequency-aware prompts~\cite{cui2024adair} or more complex designs requiring additional datasets~\cite{dudhane2024dynamic}. While effective, these approaches often increase training cost and reduce efficiency.
To further enhance generalization, MoCE-IR~\cite{zamfir2025moce} explore a mixture-of-experts design, activating specialized subnetworks for different conditions. Although this improves performance, it also increases overall architectural complexity. In contrast, our work strengthens the model’s ability to capture representative degradation information without the overhead of heavy prompts or expert routing, providing a simpler and more efficient pathway for All-in-One IR.
\section{\ourmethod}
\label{sec:method}
\begin{figure*}[!t]
    \centering
    \includegraphics[width=1.0\linewidth]{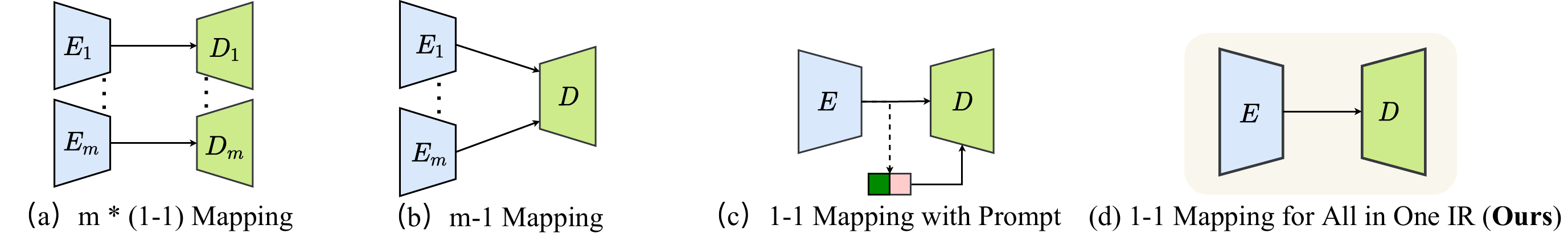}
    \caption{Mathematical formulations of mainstream paradigms for All-in-One image restoration.  
    (a) $m \times (1\!-\!1)$ mapping: each degradation type $z_i$ requires an independent encoder–decoder pair $(E_i,D_i)$.  
    (b) $m\!-\!1$ mapping with shared decoder: multiple task-specific encoders $E_i$ share a single decoder $D$.  
    (c) $1\!-\!1$ mapping with prompts: a unified encoder–decoder $(E,D)$ is conditioned on external prompts $p(z)$.  
    (d) Our proposed $1\!-\!1$ mapping: a pure unified framework where $(E,D)$ directly learns degradation-invariant yet discriminative representations, without prompts or multiple encoders.}
    \label{fig:preliminaries}
    \vspace{-3mm}
\end{figure*}

\subsection{Preliminaries}
\label{subsec:preliminary}
Formally, for the image restoration problem, given an observation:
\begin{equation}
    y = \mathcal{D}_z(x) + \epsilon,
\end{equation}
where $x$ denotes the latent clean image, $\mathcal{D}_z$ is a degradation operator parameterized by $z$ (\eg, noise, blur, haze, rain), and $\epsilon$ is an additive perturbation, the goal of IR is to recover $x$ from its degraded counterpart $y$. 

In practice, this inverse mapping is inherently ill-posed, as multiple clean images may correspond to the same degraded observation. In other words, different $x_1$ and $x_2$ may both satisfy $\mathcal{D}_z(x) + \epsilon = y$. Therefore, IR does not generally admit a mathematically unique solution. Instead of explicitly enforcing the invertibility of $\mathcal{D}_z$, modern learning-based approaches aim to estimate a statistically plausible reconstruction conditioned on $y$.

From a probabilistic perspective, the restoration model can be interpreted as a data-driven estimator of the conditional distribution $p(x \mid y)$. During training, supervision encourages convergence toward the ground-truth mode among feasible candidates, while at inference time, the prediction reflects a statistically consistent solution favored by the learned image prior rather than an arbitrary selection among multiple possibilities. This viewpoint aligns with the standard treatment of ill-posed inverse problems in the literature and explains why learning-based IR models can produce stable and perceptually coherent reconstructions even under mixed or previously unseen degradations.

This probabilistic and ill-posed nature of IR also explains why different architectural paradigms may exhibit distinct reconstruction behaviors, depending on how degradation information and prior knowledge are encoded. To situate our design within the broader landscape of All-in-One IR, we first revisit several representative architectural paradigms that differ in their degree of parameter sharing, conditioning strategy, and task dependence. This comparison provides the conceptual motivation for our formulation and clarifies the design choices made in our framework. As summarized in Fig.~\ref{fig:preliminaries}, these approaches can be interpreted as mappings from a degraded observation $y$ to a restored image $\hat{x}$, with different assumptions about how degradation type is modeled or encoded in the network~\cite{jiang2025survey}.

\begin{itemize}
    \item[(a)] \textbf{$m \times (1\!-\!1)$ Mapping.}
    Each degradation type $z_i \in \{1,\ldots,m\}$ is handled by an individual encoder–decoder pair $(E_i,D_i)$~\cite{li2023efficient,liang2021swinir}:
    \begin{equation}
        \hat{x}_i = D_i(E_i(y)), \quad i = 1,\ldots,m.
    \end{equation}
    This strategy yields strong task specialization but scales linearly with the number of degradations and prevents knowledge sharing across tasks.

    \item[(b)] \textbf{$m\!-\!1$ Mapping with Shared Decoder.}
    Multiple task-specific encoders are retained, while a shared decoder $D$ reconstructs the output~\cite{li2020all}:
    \begin{equation}
        \hat{x}_i = D(E_i(y)), \quad i = 1,\ldots,m.
    \end{equation}
    This partially amortizes parameters through a common reconstruction head, yet the storage and computation of $m$ encoders still limit efficiency and scalability.

    \item[(c)] \textbf{$1\!-\!1$ Mapping with Prompts.}
    A single encoder–decoder pair $(E,D)$ is conditioned on an external prompt $p(z)$ describing the degradation~\cite{li2023prompt,zhang2024perceive,li2023prompt}:
    \begin{equation}
        \hat{x} = D(E(y), p(z)),
    \end{equation}
    where $p(z)$ may correspond to learned tokens or textual embeddings.
    This formulation improves flexibility and controllability across degradations, but introduces auxiliary conditioning modules and requires careful prompt modeling and tuning.

    \item[(d)] \textbf{$1\!-\!1$ Mapping for All-in-One IR (adopted by our solution).}
    We advocate a pure one-to-one mapping, where a single encoder–decoder directly captures both degradation-sensitive cues and degradation-invariant structures~\cite{cui2024adair,tang2025degradation}:
    \begin{equation}
        \hat{x} = D(E(y)).
    \end{equation}
    Unlike prompt-based formulations, no explicit task tokens or routing mechanisms are introduced; instead, the network architecture is designed to enable implicit disentanglement within the shared representation space, favoring parameter efficiency and generalization across degradations.
\end{itemize}

This formulation-centric perspective highlights the trade-offs between specialization, conditioning overhead, and parameter sharing in prior All-in-One IR systems, and motivates our design choice of a unified $1\!-\!1$ mapping that seeks to balance representational robustness with computational efficiency.

\begin{figure*}[!t]
    \centering
    \includegraphics[width=1.0\linewidth]{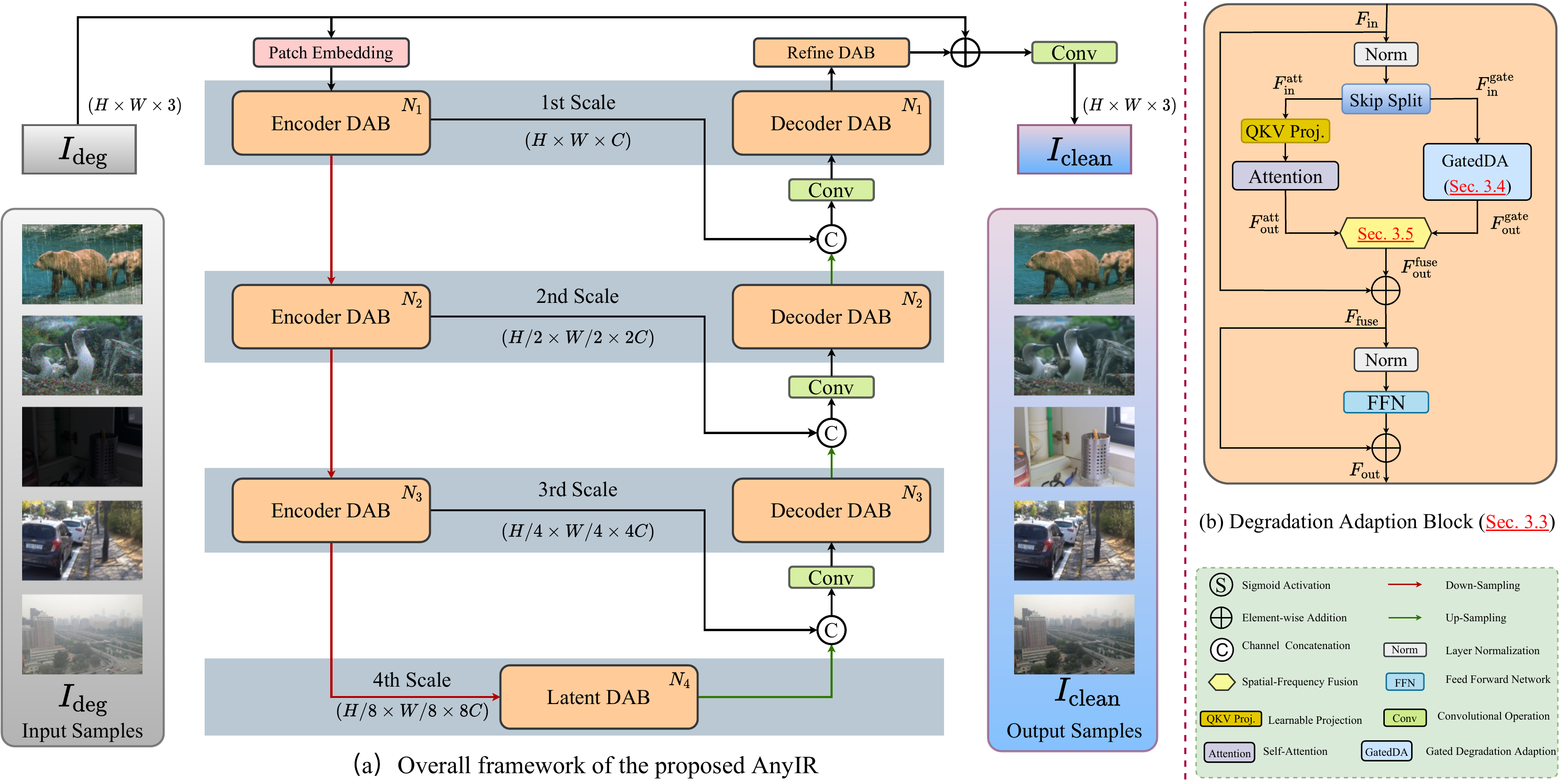}
    \caption{(a) Framework of the proposed {AnyIR}: \ie, a convolutional patch embedding, a U-shape encoder-decoder main body, and an extra refined block. (b) Structure of degradation adaptation block (DAB).
    }
    \label{fig:framework}
\end{figure*}

\subsection{Overall Framework}
\label{subsec:overall}
Building on the design space reviewed in Sec.~\ref{subsec:preliminary}, we now present our efficient All-in-One IR method, termed \ourmethod. 
As motivated by Fig.~\ref{fig:preliminaries}(d), our goal is to realize a pure $1\!-\!1$ mapping that avoids multiple encoders or external prompts, while still capturing degradation-specific cues and invariant structures within a single unified framework. 

The overall architecture of \ourmethod is illustrated in Fig.~\ref{fig:framework}. 
At a macro level, it adopts a U-shaped network~\cite{ronneberger2015unet} with four hierarchical levels. 
Each level incorporates $N_{i}, i \in [1,2,3,4]$ instances of our proposed \emph{degradation adaptation block} (DAB, Sec.~\ref{subsec:adab}), where each DAB is composed of the gated degradation adaptation module (GatedDA, Sec.~\ref{subsec:gatedDA}) and the spatial–frequency fusion algorithm (Sec.~\ref{subsec:sf_fusion}). 
Initially, a convolutional layer extracts shallow features from the degraded input, creating a patch embedding of size $H \times W \times C$. 
As in standard U-Nets, each encoder stage doubles the embedding dimension and halves the spatial resolution, with skip connections transferring information to the corresponding decoder stage. 
In the decoder, features are merged with the previous decoding stage via linear projection. 
Finally, a global skip connection links input to output, preserving high-frequency details and producing the restored image.

\subsection{Degradation Adaptation Block}
\label{subsec:adab}
The proposed degradation adaptation block (DAB) serves as the fundamental unit of \ourmethod, and its structure is shown in Fig.~\ref{fig:framework}(b). 
The design principle is to decouple global and local processing in a parameter-efficient manner, while still retaining rich feature diversity. 
Given an input feature $F_{\text{in}} \in \mathbb{R}^{H \times W \times C}$, we employ a selective channel-wise partitioning strategy:
\begin{equation}
\begin{aligned}
    F_{\text{in}}^{\text{att}} &= \{ F_{\text{in}}^{(2k-1)} \mid k \in \mathbb{Z}^+, k \leq \tfrac{C}{2} \}, \\
    F_{\text{in}}^{\text{gate}} &= \{ F_{\text{in}}^{(2k)} \mid k \in \mathbb{Z}^+, k \leq \tfrac{C}{2} \},
\end{aligned}
\label{eqn:1}
\end{equation}
where $F_{\text{in}}^{\text{att}}$ and $F_{\text{in}}^{\text{gate}} \in \mathbb{R}^{H \times W \times \tfrac{C}{2}}$ denote two interleaved channel groups. 
This skip-split partitioning is not intended to create two isolated feature branches; instead, it enforces controlled interaction across partial channel groups, so that degradation-variant cues can be emphasized while degradation-invariant structural information remains accessible within the same shared representation space.
This yields two complementary pathways: the attention branch $F_{\text{in}}^{\text{att}}$ focuses on long-range dependency modeling, while the gated branch $F_{\text{in}}^{\text{gate}}$ specializes in degradation-sensitive local adaptation. 
Compared with conventional half-split operations, the skip-split design (i) reduces the effective dimensionality per branch, lowering the complexity of attention layers~\cite{dosovitskiy2020image}, and (ii) preserves feature diversity by uniformly sampling channels, thereby avoiding information loss. 
A detailed analysis is provided in Sec.~\ref{sec:discuss}, and the visual illustration is shown in the appendix.

\begin{figure}[!t]
    \centering
    \includegraphics[width=0.8\linewidth]{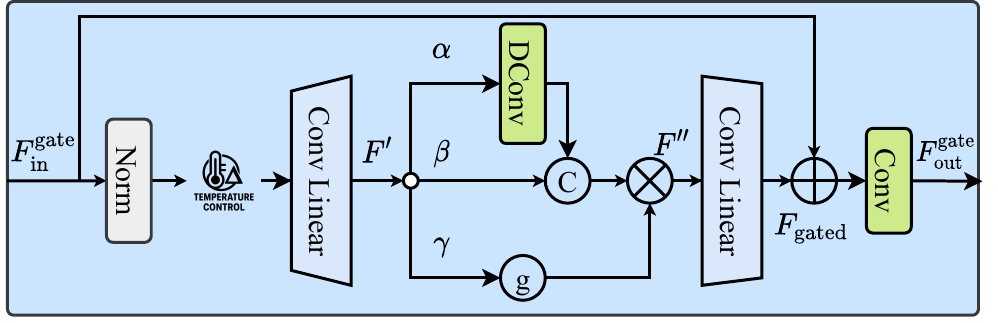}
    \caption{Structure of our GatedDA. $\oplus$, $\textcircled{c}$, $\textcircled{g}$, and $\otimes$ denote the element-wise addition, channel-wise concatenation, GELU~\cite{hendrycks2016gaussian} activation, and element-wise multiplication, respectively.}
    \label{fig:GatedDA}
\end{figure}

To capture the complex dependencies inherent to global degradation, we employ a multi-depth convolution head attention mechanism~\cite{zamir2022restormer,potlapalli2023promptir} on the feature subset \( F_{\text{in}}^{\text{att}} \), resulting in \( F_{\text{out}}^{\text{att}} \). This attention approach is particularly effective for image restoration, where degradation is often non-uniform and shaped by various intricate factors.
Specifically, \( F_{\text{in}}^{\text{att}} \) is transformed into Query (\( Q \)), Key (\( K \)), and Value (\( V \)) matrices, defined as:
$
Q = F_{\text{in}}^{\text{att}} \mathbf{W}_{\text{qry}}, \quad K = F_{\text{in}}^{\text{att}} \mathbf{W}_{\text{key}}, \quad V = F_{\text{in}}^{\text{att}} \mathbf{W}_{\text{val}},
$
where \( \mathbf{W}_{\text{qry}} \), \( \mathbf{W}_{\text{key}} \), and \( \mathbf{W}_{\text{val}} \) are learnable weights. The output \( F_{\text{out}}^{\text{att}} \) is then calculated as:
\begin{equation}
    F_{\text{out}}^{\text{att}} = \sum_i \texttt{Softmax} \left( \frac{QK^{\top}}{\sqrt{d}} \right)_i V_{i,j},
\label{eqn:attention_output}
\end{equation}
where \(\sqrt{d}\) is a scaling factor to normalize attention scores, stabilize gradients, and facilitate convergence. This attention mechanism excels at modeling long-range dependencies throughout the feature space, a vital capability for image restoration where degradation, such as noise, blur, and artifacts, exhibits spatial correlations but varies across the image~\cite{liang2021swinir,zamir2022restormer,li2023efficient}. 

In parallel, the gated branch $F_{\text{in}}^{\text{gate}}$ is processed by the GatedDA module (Sec.~\ref{subsec:gatedDA}), yielding $F_{\text{out}}^{\text{gate}}$. 
This path complements attention by enhancing localized, degradation-aware features~\cite{liang2021swinir}. Thus, the DAB combines the strengths of global modeling (attention) and local selectivity (gated convolution).

To unify both pathways, we adopt a spatial–frequency fusion strategy (Alg.~\ref{alg:sf_fusion}), producing the fused representation $F_{\text{out}}^{\text{fuse}}$. 
Finally, layer normalization, a feed-forward network (FFN), and a residual connection are applied:
\begin{equation}
    F_{\text{out}} = \operatorname{FFN}(\operatorname{Norm}(F_{\text{fuse}})) + F_{\text{fuse}}.
\end{equation}
This sequence stabilizes feature statistics and enhances expressivity, yielding a balanced representation that integrates global context with fine-grained degradation cues for high-fidelity restoration.

\subsection{Gated Degradation Adaption}
\label{subsec:gatedDA}
\begin{algorithm}[!t]
\caption{Gated Degradation Adaption}
\label{alg:gated_degradation_adaption}
\begin{algorithmic}[1]
    \Require Input $F_{\text{in}}^{\text{gate}} \in \mathbb{R}^{H \times W \times C}$, initial temperature $\tau$
    \Ensure Output $F_{\text{out}}^{\text{gate}} \in \mathbb{R}^{H \times W \times C}$
    
    \State $\hat{F} \gets \text{Norm}(F_{\text{in}}^{\text{gate}})$ \hfill \textcolor{gray}{// Normalize}
    \State $\mu \gets \frac{1}{HW} \sum_{h=1}^{H}\sum_{w=1}^{W} \hat{F}_{:, :, h, w}$ \hfill \textcolor{gray}{// per-channel mean}
    \State $\sigma \gets \sqrt{\frac{1}{HW} \sum_{h=1}^{H}\sum_{w=1}^{W} (\hat{F}_{:, :, h, w}-\mu)^2 + \delta}$ \hfill \textcolor{gray}{// per-channel std, $\delta$: small constant}
    
    \State $\Delta \gets \text{Sigmoid}(\mu + \sigma)$ \hfill \textcolor{gray}{// Temp adjust}
    \State $\tau_{\text{adj}} \gets \tau \cdot \Delta$
    
    \State $F' \gets \hat{F} \mathbf{W}_{\text{exp}}$ \hfill \textcolor{gray}{// Channel expand}
    \State $\gamma, \beta, \alpha \gets \text{split}(F')$ \hfill \hfill \textcolor{gray}{// Split $\gamma$, $\beta$, $\alpha$}
    \State $\alpha' \gets (\alpha \mathbf{W}_{\text{depth}}) \cdot (1 + \tau_{\text{adj}}$) \hfill \textcolor{gray}{// Depthwise conv}
    
    \State $F_{\text{gated}} \gets \sigma(\gamma) \cdot \text{concat}(\beta, \alpha') \mathbf{W}_{\text{gate}}$ \hfill \textcolor{gray}{// Gate combine}
    \State $F_{\text{out}}^{\text{gate}} \gets (F_{\text{gated}} + F_{\text{in}}^{\text{gate}}) \mathbf{W}_{\text{proj}}$ \hfill \textcolor{gray}{// Residual, proj}
    
    \State \Return $F_{\text{out}}^{\text{gate}}$
\end{algorithmic}
\end{algorithm}
To capture local degradation-aware details, we leverage the selective properties of gated convolution, forming the GatedDA (Fig.~\ref{fig:GatedDA}). 
Given an input feature \( F_{\text{in}}^{\text{gate}} \in \mathbb{R}^{H \times W \times C}\), a layer normalization is first applied to stabilize the feature distributions, followed by a 1\(\times\)1 convolution that expands the channel dimension to \( \text{hidden} = r_{\text{expan}} \cdot C \), where \( r_{\text{expan}} \) is the expansion ratio, as shown in Alg.~\ref{alg:gated_degradation_adaption}. 
To adaptively respond to varying intensities, we introduce a temperature adjustment mechanism. 
Based on the input’s \textit{mean} and \textit{standard} deviation, the initial temperature \( \tau \) is modulated as \( \tau_{\text{adj}} = \tau \cdot \Delta \), where \( \Delta \) serves as a dynamic scaling factor (steps 2–5 of Alg.~\ref{alg:gated_degradation_adaption}). 
This design is motivated by the fact that many degradations exhibit spatially non-uniform characteristics; therefore, rather than applying a single global transformation, the gated module performs content- and region-aware modulation on features, enabling localized adaptation to different degradation strengths.
This adjustment improves the model’s sensitivity to subtle degradation patterns, promoting detailed feature capture.

The expanded feature \( F' \) ($\mathbf{W}_{\text{exp}}$ denotes the 1\(\times\)1 convolution weights used for channel expansion) is then split into three parts along channels: 
\(\alpha\), \(\beta\), and \(\gamma\) (as shown in step 7, which correspond to the \textit{scaled}, \textit{ego}, and \textit{shifted} features). 
{Specifically, we take (\nicefrac{hidden}{4}, \nicefrac{hidden}{4}, \nicefrac{hidden}{2}) for (\(\alpha\), \(\beta\), \(\gamma\)) in this paper.}
Here, \(\alpha\) undergoes depthwise convolution with \( \tau_{\text{adj}} \) to capture spatial details, \(\beta\) retains the original information, and \(\gamma\) is activated with GELU~\cite{hendrycks2016gaussian} to enable a non-linear gated selection mechanism. 
Specifically, these components are recombined and modulated, selectively emphasizing critical features (Step 9 in Alg.~\ref{alg:gated_degradation_adaption}).
Finally, \( F'' \) is projected back to the original channel dimension and combined with \( F_{in}^{\text{gate}} \) through a skip connection to prevent loss of information. 
A final 1\(\times\)1 convolution fuses the degradation-adapted features with the input, resulting in the output \( F_{\text{out}}^{\text{gate}} \) (Steps 10–11 in Alg.~\ref{alg:gated_degradation_adaption}). 
Please note that $\mathbf{W}_{\text{depth}}$, $\mathbf{W}_{\text{gate}}$, and $\mathbf{W}_{\text{proj}}$ denote the learnable convolution weights for depth-wise convolution, gated convolution, and projection convolution operations.

Owing to its design, GatedDA dynamically adjusts its internal temperature and gating behavior, enabling the network to capture degradation-aware features adaptively. This makes it a natural complement to the global attention branch, providing localized detail enhancement and stronger robustness to spatially varying degradations. More analysis is provided in the appendix.

From an efficiency perspective, the proposed design improves computational economy at the architectural level rather than relying on model scaling. First, each Degradation Adaptation Block partitions the feature channels into two subsets, where only half of the channels are processed by global attention. Since our attention operator follows the Restormer-style channel-wise formulation whose complexity grows quadratically with spatial resolution, reducing the attention channels by half directly lowers the computational cost of the attention pathway while preserving its ability to model long-range dependencies. Second, the remaining channels are processed by the proposed GatedDA module, which enhances the representational capacity of the block through lightweight, convolution-based local adaptation; this branch has substantially lower complexity than applying attention on the same number of channels, further contributing to overall efficiency. Finally, because each block combines a stronger per-block representation with a more balanced global–local decomposition, the backbone can be instantiated with a smaller U-shaped hierarchy (\eg, [3, 5, 5, 7] instead of the deeper [4, 6, 6, 8] configurations used in prior All-in-One IR methods such as PromptIR~\cite{potlapalli2023promptir} and MoCE-IR~\cite{zamfir2025moce}), while still achieving superior restoration accuracy.

\subsection{Spatial-Frequency Fusion Algorithm}
\label{subsec:sf_fusion}
\setlength{\textfloatsep}{7pt}
\begin{algorithm}[t]
\caption{Spatial-Frequency Fusion}
\label{alg:sf_fusion}
\begin{algorithmic}[1]
    \Require $F_{\text{att}}$, $F_{\text{gate}}$, fusion weight $\lambda$
    \Ensure Fused output $F_{\text{fuse}}$

    \Statex \textbf{Main Procedure:}
    \State $F_s \gets \Call{SpatialFusion}{F_{\text{att}}, F_{\text{gate}}}$ \hfill \textcolor{gray}{// spatial cross-enhance}
    \State $F_f \gets \Call{FrequencyFusion}{F_{\text{att}}, F_{\text{gate}}}$ \hfill \textcolor{gray}{// detail-enhance}
    \State $F_{\text{fuse}} \gets \lambda \cdot F_s + (1 - \lambda) \cdot F_f$ \hfill \textcolor{gray}{// weighted fusion}
    \State \Return $F_{\text{fuse}}$

    \vspace{0.1em}
    \Statex \textbf{Function: SpatialFusion}
    \Function{SpatialFusion}{$F_{\text{att}}, F_{\text{gate}}$}
        \State $F^{\text{ag}} \gets F_{\text{att}} + \text{Sigmoid}(F_{\text{gate}})$ \hfill \textcolor{gray}{// gate-enhanced attention}
        \State $F^{\text{ga}} \gets F_{\text{gate}} + \text{Sigmoid}(F_{\text{att}})$ \hfill \textcolor{gray}{// attention-enhanced gate}
        \State \Return $\text{Concat}(F^{\text{ag}}, F^{\text{ga}})$ \hfill \textcolor{gray}{// channel-wise merge}
    \EndFunction

    \vspace{0.1em}
    \Statex \textbf{Function: FrequencyFusion}
    \Function{FrequencyFusion}{$F_{\text{att}}, F_{\text{gate}}$}
        \State $\hat{F}_{\text{att}} \gets \text{rfft}_{2D}(F_{\text{att}})$, $\hat{F}_{\text{gate}} \gets \text{rfft}_{2D}(F_{\text{gate}})$ \hfill \textcolor{gray}{// real FFT}
        \State $\hat{F} \gets \hat{F}_{\text{att}} + \hat{F}_{\text{gate}}$ \hfill \textcolor{gray}{// freq combine}
        \State $F_{\text{freq}} \gets \text{irfft}_{2D}(\hat{F})$ \hfill \textcolor{gray}{// inverse FFT}
        \State \Return $\text{Repeat}(F_{\text{freq}})$ \hfill \textcolor{gray}{// match channels}
    \EndFunction  
\end{algorithmic}
\end{algorithm}
To enable effective interaction between contextualized global attention and
degradation-sensitive local GatedDA features, we introduce a spatial–frequency fusion algorithm (\textit{that is}, Alg.~\ref{alg:sf_fusion}).
The fusion is carried out in two complementary domains: spatial and frequency.

In the spatial branch, we apply a cross-enhancement mechanism (\ie, SpationFusion in Alg.~\ref{alg:sf_fusion}) to refine the attention feature $F_{att}$ and gated feature $F_{gate}$ mutually. Each branch is modulated by the sigmoid-activated signal from the other, enabling a dynamic message passing across representations. The enhanced features are then concatenated along the channel dimension to form the spatial representation.

In parallel, we perform a lightweight frequency-domain fusion to capture structural alignment. \(F_{att}\) and \(F_{gate}\) are first transformed using real-valued 2D Fast Fourier Transform (\ie, $\text{rfft}_{2D}(*)$) in Step-11 of Alg.~\ref{alg:sf_fusion}, additively combined in the frequency domain (Step-12), and then projected back via inverse FFT (\ie, $\text{irfft}_{2D}(*)$) (Step-13). The output is duplicated (\ie, $\text{Repeat}(*)$) along the channel to match the spatial counterpart (Step-14). 
This design leverages the complementary roles of the two domains: the spatial branch emphasizes local structural fidelity, whereas the frequency branch stabilizes global statistics and degradation patterns, which is particularly beneficial under mixed or compound degradations.

The final fused representation is obtained by a weighted summation of the two branches (Step-3 of Alg.~\ref{alg:sf_fusion}), controlled by a scalar $\lambda$. A convolutional projection with learnable weights $\mathbf{W}_{\text{fuse}}$ is then applied, followed by a residual connection to the original input:
\begin{equation}
    F_{\text{out}}^{\text{fuse}} \gets \operatorname{Conv}_{\mathbf{W}_{\text{fuse}}}(F_{\text{fuse}}) + F_{\text{in}}.
\end{equation}

This fusion strategy leverages the complementary strengths of attention-based global context and gated local priors from both spatial and frequency domains, producing a rich and adaptive representation for downstream restoration. Notably, the design aligns with signal-domain interpretability while enhancing generalization across diverse degradation types.
\section{Experimental Results}
\label{sec:experiments}
\begin{figure*}[t]
    \centering
    \includegraphics[width=0.99\linewidth]{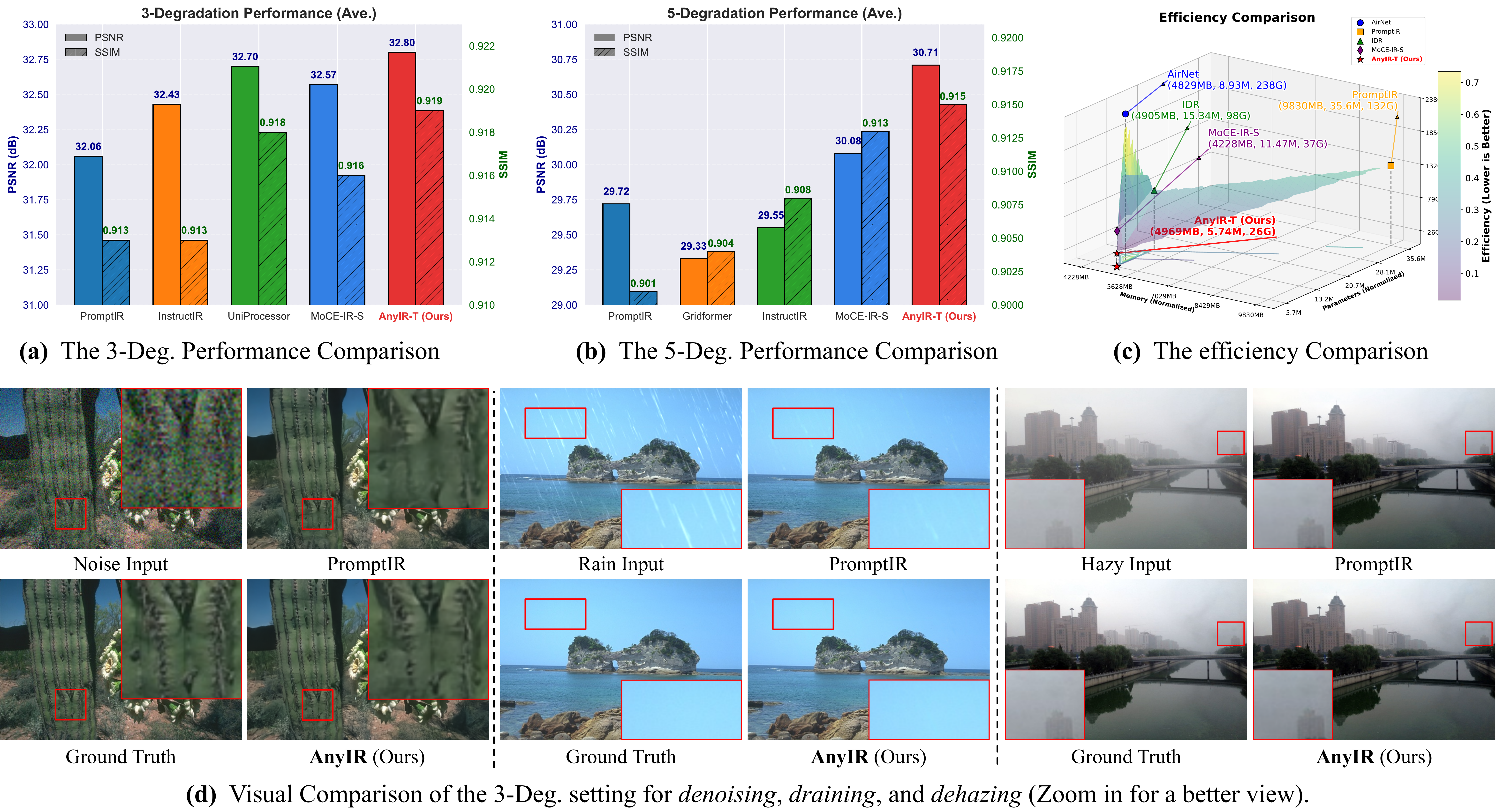}
    \caption{
    Overall quantitative performance (3-Degradation \& 5-Degradation settings), efficiency comparison, and the qualitative comparison ( Zoom in for a better view).
    }
    \label{fig:teaser}
\end{figure*}

We conduct experiments adhering to the protocols of previous general IR works \cite{potlapalli2023promptir,zhang2023ingredient} in three main settings: 
\textit{(1) All-in-One (3Degrations)}, 
\textit{(2) All-in-One (5Degrations)}, 
\textit{(3) Mix-Degradation Setting}, and
\textit{(4) Zero-Shot Unseen Setting}
More experimental details and the introduction of the data set are provided in our \textit{Supp. Mat.}

Before presenting the detailed results for each setting, we first provide an
overview comparison to summarize the overall performance, efficiency, and
visual restoration quality of \ourmethod{} against representative All-in-one IR
methods, as shown in Fig.~\ref{fig:teaser}.

\subsection{State of the Art Comparisons}
\begin{table*}[t]
    \centering
    \scriptsize
    \fboxsep0.75pt
    \setlength\tabcolsep{1pt}
    \setlength{\extrarowheight}{0.8pt}
    \caption{\textit{Comparison to state-of-the-art on three degradations.} PSNR (dB, $\uparrow$) and SSIM ($\uparrow$) metrics are reported on the full RGB images. \textcolor{tabred}{\textbf{Best}} and \textcolor{tabblue}{\textbf{Second Best}} performances is highlighted. Our method sets a new state-of-the-art on average across all benchmarks while being significantly more efficient than prior work. ‘-’ represents unreported results.}
    \vspace{-3mm}
    \label{tab:exp:3deg}
    \begin{tabularx}{\textwidth}{p{5cm}*{15}{c}}
    \toprule
     \multirow{2}{*}{Method} & \multirow{2}{*}{Venue.} 
     & \multirow{2}{*}{Params.} & \multicolumn{2}{c}{\textit{Dehazing}} & \multicolumn{2}{c}{\textit{Deraining}} & \multicolumn{6}{c}{\textit{Denoising}} 
     & \multicolumn{2}{c}{\multirow{2}{*}{Average}}  \\
     \cmidrule(lr){4-5} \cmidrule(lr){6-7} \cmidrule(lr){8-13} 
     &&& \multicolumn{2}{c}{SOTS} & \multicolumn{2}{c}{Rain100L} & \multicolumn{2}{c}{BSD68\textsubscript{$\sigma$=15}} & \multicolumn{2}{c}{BSD68\textsubscript{$\sigma$=25}} & \multicolumn{2}{c}{BSD68\textsubscript{$\sigma$=50}} &  \\
     \midrule
        \rowcolor{gray!10} BRDNet~\cite{tian2000brdnet} & NN'20 &- & 23.23 & {.895} & 27.42 & {.895} & 32.26 & {.898} & 29.76 & {.836} & 26.34 & {.693} & 27.80 & {.843} \\
        LPNet~\cite{gao2019dynamic} & CVPR'19 &- & 20.84 & {.828} & 24.88 & {.784} & 26.47 & {.778} & 24.77 & {.748} & 21.26 & {.552} & 23.64 & {.738} \\
        \rowcolor{gray!10} FDGAN~\cite{dong2020fdgan}  & AAAI'20&- & 24.71 & {.929} & 29.89 & {.933} & 30.25 & {.910} & 28.81 & {.868} & 26.43 & {.776} & 28.02 & {.883} \\
        DL~\cite{fan2019dl} & TPAMI'19 &2M & 26.92 & {.931} & 32.62 & {.931} & 33.05 & {.914} & 30.41 & {.861} & 26.90 & {.740}  & 29.98 & {.876}\\
        \rowcolor{gray!10} 
        MPRNet~\cite{zamir2021multi} & CVPR'21 & 16M & 25.28 & {.955} & 33.57 & {.954} & 33.54 & {.927} & 30.89 & {.880} & 27.56 & {.779} & 30.17 & {.899} \\
        AirNet~\cite{li2022all} & CVPR'22 & 9M & 27.94 & {.962} & 34.90 & {.967} & 33.92 & {.933} & 31.26 & {.888} & 28.00 & {.797} & 31.20 & {.910} \\
        \rowcolor{gray!10} 
        NDR~\cite{yao2024neural} & TIP'24 & 28.4M & 25.01 & {.860} & 28.62 & {.848} & 28.72 & {.826} & 27.88 & {.798} & 26.18 & {.720} & 25.01 & {.810} \\
        PromptIR~\cite{potlapalli2023promptir} & NeurIPS'23 & 36M  & 30.58 & .974 & 36.37 &  .972 & 33.98 &  \textcolor{tabblue}{\textbf{.933}} & {31.31} & \textcolor{tabblue}{\textbf{.888}} & {28.06} &  {.799} & {32.06} &  {.913} \\
        \rowcolor{gray!10} 
        MoCE-IR-S~\cite{zamfir2025moce} & CVPR'25 & 11M  & 30.98 & \textcolor{tabblue}{\textbf{.979}} & {38.22} &  \textcolor{tabblue}{\textbf{.983}} & 34.08 &  \textcolor{tabblue}{\textbf{.933}} & {31.42} & \textcolor{tabblue}{\textbf{.888}} & {28.16} &  {.798} & {32.57} &  {.916} \\
        \midrule
        \ourmethod-T(\textit{Ours}) & 2025 & 6M & \textcolor{tabblue}{\textbf{31.70}} &  \textcolor{tabred}{\textbf{.982}} & \textcolor{tabblue}{\textbf{38.51}} &  \textcolor{tabred}{\textbf{.983}} & \textcolor{tabblue}{\textbf{34.12}} &  \textcolor{tabred}{\textbf{.936}} & \textcolor{tabblue}{\textbf{31.46}} & \textcolor{tabred}{\textbf{.893}} & \textcolor{tabblue}{\textbf{28.20}} & \textcolor{tabblue}{\textbf{.804}} & \textcolor{tabblue}{\textbf{32.80}} &  \textcolor{tabblue}{\textbf{.919}} \\
        \ourmethod-S(\textit{Ours}) & 2025 & 9M & \textcolor{tabred}{\textbf{31.85}} &  \textcolor{tabred}{\textbf{.982}} & \textcolor{tabred}{\textbf{38.56}} &  \textcolor{tabred}{\textbf{.983}} & \textcolor{tabred}{\textbf{34.15}} &  \textcolor{tabred}{\textbf{.936}} & \textcolor{tabred}{\textbf{31.49}} & \textcolor{tabred}{\textbf{.893}} & \textcolor{tabred}{\textbf{28.24}} & \textcolor{tabred}{\textbf{.806}} & \textcolor{tabred}{\textbf{32.86}} &  \textcolor{tabred}{\textbf{.920}} \\
        \midrule
        \multicolumn{14}{c}{Methods with the assistance of vision language, multi-task learning, natural language prompts, or multi-modal control} \\
        \midrule 
        \rowcolor{gray!10} DA-CLIP~\cite{luo2023controlling} & ICLR'24& 125M & 29.46 & .963 & 36.28 & .968 &  30.02 & .821 & 24.86 & .585 & 22.29 & .476 & - & - \\
        Art$_{PromptIR}$~\cite{wu2024harmony} &ACM MM'24 & 36M & 30.83 &.979 & 37.94 & .982 &34.06 & .934 & 31.42 & .891 & 28.14 & .801 & 32.49 & .917 \\
        \rowcolor{gray!10} InstructIR-3D~\cite{conde2024high} & ECCV'24 & 16M & 30.22 &.959 & 37.98 & .978 & 34.15& .933 &31.52&.890& 28.30&.804 &32.43&.913 \\
        UniProcessor~\cite{duan2025uniprocessor} &ECCV'24& 1002M & 31.66 & .979 &  38.17 & .982 & 34.08 &.935 &  31.42 & .891 & 28.17 & .803 & 32.70 & .918  \\
    \bottomrule
    \end{tabularx}
\end{table*}

\begin{table*}[t]
    \centering
    \scriptsize
    \fboxsep0.75pt
    \setlength\tabcolsep{1pt}
    \setlength{\extrarowheight}{0.2pt}
    \caption{\textit{Comparison to state-of-the-art on five degradations.} PSNR (dB, $\uparrow$) and SSIM ($\uparrow$) metrics are reported on the full RGB images with $(^\ast)$ denoting general image restorers, others are specialized all-in-one approaches. \textcolor{tabred}{\textbf{Best}} and \textcolor{tabblue}{\textbf{Second Best}} performances is highlighted.}
    \vspace{-2mm}
    \label{tab:exp:5deg}
    \begin{tabularx}{\textwidth}{p{5cm}*{15}{c}}
    \toprule
    \multirow{2}{*}{Method} &\multirow{2}{*}{Venue}& \multirow{2}{*}{Params.} 
    & \multicolumn{2}{c}{\textit{Dehazing}} & \multicolumn{2}{c}{\textit{Deraining}} & \multicolumn{2}{c}{\textit{Denoising}} 
    & \multicolumn{2}{c}{\textit{Deblurring}} & \multicolumn{2}{c}{\textit{Low-Light}} & \multicolumn{2}{c}{\multirow{2}{*}{Average}}  \\
    \cmidrule(lr){4-5} \cmidrule(lr){6-7} \cmidrule(lr){8-9} \cmidrule(lr){10-11} \cmidrule(lr){12-13}
    &&& \multicolumn{2}{c}{SOTS} & \multicolumn{2}{c}{Rain100L} & \multicolumn{2}{c}{BSD68\textsubscript{$\sigma$=25}} 
    & \multicolumn{2}{c}{GoPro} & \multicolumn{2}{c}{LOLv1} &  \\
    \midrule
    
    \rowcolor{gray!10} NAFNet$^\ast$~\cite{chen2022simple} & ECCV'22 & 17M & 25.23 & {.939} & 35.56 & {.967} & 31.02 & {.883} & 26.53 & {.808} & 20.49 & {.809} & 27.76 & {.881} \\
    DGUNet$^\ast$~\cite{mou2022deep} & CVPR'22& 17M & 24.78 & {.940} & 36.62 & {.971} & 31.10 & {.883} & 27.25 & {.837} & 21.87 & {.823} & 28.32 & {.891} \\
    \rowcolor{gray!10} SwinIR$^\ast$~\cite{liang2021swinir} &ICCVW'21& 1M & 21.50 & {.891} & 30.78 & {.923} & 30.59 & {.868} & 24.52 & {.773} & 17.81 & {.723} & 25.04 & {.835} \\ 
    Restormer$^\ast$~\cite{zamir2022restormer} &CVPR'22& 26M & 24.09 & {.927} & 34.81 & {.962} & 31.49 & {.884} & 27.22 & {.829} & 20.41 & {.806} & 27.60 & {.881} \\
    \rowcolor{gray!10} MambaIR$^\ast$~\cite{guo2024mambair} &ECCV'24& 27M & 25.81 & .944 & 36.55 & .971 & 31.41 & .884 & 28.61 & .875 & 22.49 & .832 & 28.97 &.901 \\
    \midrule
    DL~\cite{fan2019dl} & TPAMI'19 &2M & 20.54 & {.826} & 21.96 & {.762} & 23.09 & {.745} & 19.86 & {.672} & 19.83 & {.712} & 21.05 & {.743} \\
    \rowcolor{gray!10} TransWeather~\cite{valanarasu2022transweather} & CVPR'22 & 38M & 21.32 & {.885} & 29.43 & {.905} & 29.00 & {.841} & 25.12 & {.757} & 21.21 & {.792} & 25.22 & {.836} \\
    TAPE~\cite{liu2022tape} &ECCV'22& 1M & 22.16 & {.861} & 29.67 & {.904} & 30.18 & {.855} & 24.47 & {.763} & 18.97 & {.621} & 25.09 & {.801} \\
    \rowcolor{gray!10} 
    AirNet~\cite{li2022all} &CVPR'22& 9M & 21.04 & {.884} & 32.98 & {.951} & 30.91 & {.882} & 24.35 & {.781} & 18.18 & {.735} & 25.49 & {.847} \\
    IDR~\cite{zhang2023ingredient} &CVPR'23& 15M & {25.24} & {.943} & {35.63} & {.965} & 31.60 & \textcolor{tabblue}{\textbf{.887}} & {27.87} & {.846} & 21.34 & {.826} & {28.34} & {.893} \\
    \rowcolor{gray!10}
    PromptIR~\cite{potlapalli2023promptir} &NeurIPS'23& 36M & 30.41 & {.972} & 36.17 & {.970} & 31.20 & {.885} & 27.93 & {.851} & \textcolor{tabblue}{\textbf{22.89}} & {.829} & 29.72 & {.901} \\
    MoCE-IR-S~\cite{zamfir2025moce} & CVPR'25 & 11M & 31.33 & {.978} & 37.21 & {.978} & {31.25} & {.884} & \textcolor{tabred}{\textbf{28.90}} & \textcolor{tabred}{\textbf{.877}} & {21.68} & {.851} & 30.08 & {.913} \\
    \midrule
    \ourmethod-T (\textit{ours}) &2025& 6M & \textcolor{tabblue}{\textbf{31.50}} & \textcolor{tabblue}{\textbf{.981}} & \textcolor{tabblue}{\textbf{38.81}} & \textcolor{tabblue}{\textbf{.984}} & \textcolor{tabblue}{\textbf{31.40}} & 
    \textcolor{tabred}{\textbf{.892}} & 
    28.35 & {.863} & {22.68} & \textcolor{tabblue}{\textbf{.854}} & \textcolor{tabblue}{\textbf{30.71}} & \textcolor{tabblue}{\textbf{.915}} \\
    \ourmethod-S (\textit{ours}) &2025& 9M & \textcolor{tabred}{\textbf{31.77}} & \textcolor{tabred}{\textbf{.982}} & \textcolor{tabred}{\textbf{39.00}} & \textcolor{tabred}{\textbf{.983}} & \textcolor{tabred}{\textbf{31.44}} & 
    \textcolor{tabred}{\textbf{.892}} & 
    \textcolor{tabblue}{\textbf{28.52}} & 
    \textcolor{tabblue}{\textbf{.867}} & \textcolor{tabred}{\textbf{23.03}} & \textcolor{tabred}{\textbf{.857}} & \textcolor{tabred}{\textbf{30.75}} & \textcolor{tabred}{\textbf{.916}} \\
    \midrule
    \multicolumn{14}{c}{Methods with the assistance of natural language prompts or multi-task learning} \\
    \midrule
    \rowcolor{gray!10}InstructIR-5D~\cite{conde2024high} &ECCV'24& 16M & 36.84 & .973 & 27.10 & .956 & 31.40 & .887 & 29.40 & .886 & 23.00 & .836 & 29.55 & .908 \\
    Art$_{PromptIR}$~\cite{wu2024harmony} & ACM MM'24& 36M& 29.93 & .908 & 22.09 & .891 & 29.43 & .843 & 25.61 & .776 & 21.99 & .811 & 25.81 & .846 \\
    \bottomrule
    \end{tabularx}
\end{table*}

\noindent\textbf{Three Degradations.}
We evaluate our All-in-One restorer, \ourmethod, against other specialized methods listed in Tab.~\ref{tab:exp:3deg}, all trained on three degradations: dehazing, deraining, and denoising. \ourmethod consistently outperforms all the comparison methods, even for those with the assistance of language, multi-task, or prompts. In particular, \ourmethod outperforms the baseline method PromptIR by 
\textbf{1.12dB}, \textbf{2.14dB} on dehazing and draining, and \textbf{0.74dB} on average, while maintaining \textbf{80\%} fewer parameters.

\noindent\textbf{Five Degradations.} 
Extending the three degradation tasks to include deblurring and low-light enhancement~\cite{li2022all,zhang2023ingredient}, we validate the comprehensive performance of our method in an All-in-One setting. As shown in Tab.~\ref{tab:exp:5deg}, \ourmethod effectively leverages degradation-specific features, surpassing AirNet~\cite{li2022all} and IDR~\cite{zhang2023ingredient} by an average of 5.16 dB and 2.31 dB, respectively, with 33\% and 60\% fewer parameters. Compared to the most recent method, MoCE-IR~\cite{zamfir2025moce}, besides 0.05 dB lower on deblurring, we outperform it on all the rest tasks with a 0.57dB PSNR improvement on average.

\begin{table}[t]
    \centering
    \tiny
    \setlength{\tabcolsep}{0.55pt}
    \caption{\textit{Comparison to state-of-the-art on composited degradations.} PSNR (dB, $\uparrow$) and \colorbox{clblue!50}{SSIM ($\uparrow$)} are reported on the full RGB images.
    Our method consistently outperforms even larger models, with favorable results in composited degradation scenarios.}
    \label{tab:mixdeg}
    \begin{tabularx}{\textwidth}{p{1.6cm}c*{8}{c}*{10}{c}*{4}{c}cc}
    \toprule
    \multirow{2}{*}{Method} & \multirow{2}{*}{Params.} & \multicolumn{8}{c}{\textit{CDD11-Single}} & \multicolumn{10}{c}{\textit{CDD11-Double}} & \multicolumn{4}{c}{\textit{CDD11-Triple}} & \multicolumn{2}{c}{\multirow{2}{*}{Avg.}}\\
    \cmidrule(lr){3-10} \cmidrule(lr){11-20} \cmidrule(lr){21-24}
    && \multicolumn{2}{c}{Low~(L)} & \multicolumn{2}{c}{Haze~(H)} & \multicolumn{2}{c}{Rain~(R)} & \multicolumn{2}{c}{Snow~(S)}
    & \multicolumn{2}{c}{L+H} & \multicolumn{2}{c}{L+R} & \multicolumn{2}{c}{L+S} & \multicolumn{2}{c}{H+R} & \multicolumn{2}{c}{H+S} 
    &  \multicolumn{2}{c}{L+H+R} &  \multicolumn{2}{c}{L+H+S} \\ 
    \midrule
    \rowcolor{gray!10} 
    AirNet & 9M
    & 24.83&\cc{.778} & 24.21&\cc{.951} & 26.55&\cc{.891} & 26.79&\cc{.919}
    & 23.23&\cc{.779} & 22.82&\cc{.710} & 23.29&\cc{.723} & 22.21&\cc{.868} & 23.29&\cc{.901}
    & 21.80&\cc{.708} & 22.24&\cc{.725} & 23.75&\cc{.814} \\
    PromptIR & 36M
    & 26.32&\cc{.805} & 26.10&\cc{.969} & 31.56&\cc{.946} & 31.53&\cc{.960} 
    & 24.49&\cc{.789} & 25.05&\cc{.771} & 24.51&\cc{.761} & 24.54&\cc{.924} & 23.70&\cc{.925} 
    & 23.74&\cc{.752} & 23.33&\cc{.747} & 25.90&\cc{.850} \\
    \rowcolor{gray!10} 
    WGWSNet & 26M
    & 24.39&\cc{.774} & 27.90&\cc{.982} & 33.15&\cc{.964} & 34.43&\cc{.973} 
    & 24.27&\cc{.800} & 25.06&\cc{.772} & 24.60&\cc{.765} & 27.23&\cc{.955} & 27.65&\cc{.960}  
    & 23.90&\cc{.772} & 23.97&\cc{.771} & 26.96&\cc{.863} \\
    WeatherDiff & 83M
    & 23.58&\cc{.763} & 21.99&\cc{.904} & 24.85&\cc{.885} & 24.80&\cc{.888} 
    & 21.83&\cc{.756} & 22.69&\cc{.730} & 22.12&\cc{.707} & 21.25&\cc{.868} & 21.99&\cc{.868} 
    & 21.23&\cc{.716} & 21.04&\cc{.698} & 22.49&\cc{.799} \\
    \rowcolor{gray!10} 
    OneRestore & 6M
    & 26.48&\cc{{.826}} & 32.52&\cc{.990} & 33.40&\cc{.964} & 34.31&\cc{.973}  
    & 25.79&\cc{{.822}} & 25.58&\cc{.799} & 25.19&\cc{.789} & {29.99}&\cc{.957} & {30.21}&\cc{.964} 
    & 24.78&\cc{.788} & 24.90&\cc{{.791}} & 28.47&\cc{.878} \\
    MoCE-IR & 11M 
    & {27.26}&\cc{.824} & {32.66}&\cc{{.990}} & {34.31}&\cc{{.970}} & {35.91}&\cc{{.980}}
    & {26.24}&\cc{.817} & {26.25}&\cc{{.800}} &{26.04}&\cc{{.793}} & 29.93&\cc{{.964}} & 30.19& \cc{{.970}}
    & {25.41}& \cc{{.789}} &{25.39}&\cc{.790} & {29.05} & \cc{.881} \\
    \midrule 
    AnyIR-T(\textit{ours}) & 6M & 27.40 &\cc{.833} & 33.41 & \cc{.991} & 34.53 & \cc{.970} & 36.07 & .979 & 26.53 & \cc{.827} & 26.55 & \cc{.810} & 26.36 & \cc{.802} & 30.40 & \cc{.965} & 30.65  &\cc{.970} & 25.66 & \cc{.800} & 25.86 & \cc{.801} & 29.40 & \cc{.886} \\
    AnyIR-S(\textit{ours}) & 10M & \sotaa{\textbf{27.47}} & \cc{\sotaa{\textbf{.835}}} & \sotaa{\textbf{34.38}} & \cc{\sotaa{\textbf{.992}}} & \sotaa{\textbf{34.64}} & \cc{\sotaa{\textbf{.971}}} & \sotaa{\textbf{36.21}} & \cc{\sotaa{\textbf{.981}}} & \sotaa{\textbf{26.54}} & \cc{\sotaa{\textbf{.830}}} & \sotaa{\textbf{26.49}} & \cc{\sotaa{\textbf{.812}}} & \sotaa{\textbf{26.45}} & \cc{\sotaa{\textbf{.805}}} & \sotaa{\textbf{30.98}} & \cc{\sotaa{\textbf{.967}}} & \sotaa{\textbf{31.55}} & \cc{\sotaa{\textbf{.972}}} & \sotaa{\textbf{25.82}} & \cc{\sotaa{\textbf{.804}}} & \sotaa{\textbf{26.01}} & \cc{\sotaa{\textbf{.805}}} & \sotaa{\textbf{29.69}} & \cc{\sotaa{\textbf{.889}}}  \\
    \bottomrule
\end{tabularx}
\end{table}
\noindent\textbf{Mixed Degradation.}
We conduct experiments on the CDD-11~\cite{guo2024onerestore} dataset, a challenging benchmark for mixed degradation restoration tasks, combining real-world degradations such as low-light conditions, haze, rain, and snow. 
As shown in Tab.~\ref{tab:mixdeg}, our method consistently surpasses other state-of-the-art approaches like AirNet~\cite{li2022all}, PromptIR~\cite{potlapalli2023promptir}, WGWSNet~\cite{ZhuWFYGDQH23}, WeatherDiff~\cite{weatherdiffusion}, OneRestore~\cite{guo2024onerestore}, and MoCE-IR~\cite{zamfir2025moce}.
These results demonstrate its robustness in addressing complex interactions among degradations.
The superior performance of \ourmethod validates its advanced degradation modeling and fusion mechanisms, enabling effective restoration in various scenarios.

\begin{table}[!t]
    \centering
    \footnotesize
    \caption{Unseen \textbf{Desnowing} on \textit{CSD}~\cite{chen2021all} dataset.}
    \label{tab:unseen}
    \setlength{\tabcolsep}{6pt}
    \setlength{\extrarowheight}{0.5pt}
    \begin{tabular}{lcc}
        \toprule
        Method & PSNR & SSIM \\
        \midrule
        \rowcolor{gray!10} AirNet~\cite{li2022all} & 19.32 & .733 \\
        PromptIR~\cite{potlapalli2023promptir} & 20.47 & .764 \\
        \rowcolor{gray!10}  MoceIR-S~\cite{zamfir2025moce} & 21.09 & .771 \\
        \ourmethod (Ours) & \textcolor{tabred}{\textbf{21.64}} & \textcolor{tabred}{\textbf{.787}} \\
        \bottomrule
    \end{tabular}
\end{table}

\begin{table}[!t]
    \centering
    \caption{\textit{Zero-Shot} Cross-Domain Underwater Image Enhancement Results.}
    \label{table:underwaterIR}
    \setlength{\extrarowheight}{0.1pt}
    \setlength{\tabcolsep}{6pt}
    \scalebox{0.73}{
    \begin{tabular}{l | c c }
    \toprule[.1em]
    Method & PSNR (dB, $\uparrow$) & SSIM ($\uparrow$) 	\\ 
    \midrule
    \rowcolor{gray!10} SwinIR~\cite{liang2021swinir} & 15.31 & .740 \\
    NAFNet~\cite{chu2022nafssr} & 15.42 & .744 \\	
    \rowcolor{gray!10} Restormer~\cite{zamir2022restormer} & 15.46 & .745 \\
    \midrule
    AirNet~\cite{li2022all} & 15.46 & .745 \\
    \rowcolor{gray!10} IDR~\cite{zhang2023ingredient} & 15.58 & .762 \\
    PromptIR~\cite{potlapalli2023promptir} & 15.48 & .748 \\ 
    \rowcolor{gray!10} MoCE-IR~\cite{zamfir2025moce} & 15.91 & .765 \\
    \midrule
    \ourmethod-S \textit{(Ours)} & \sotaa{\textbf{16.78}} & \sotaa{\textbf{.770}} \\
    \bottomrule[.1em]
    \end{tabular}
    }
\end{table}

\noindent\textbf{Zero-Shot Unseen Degradation.} 
To assess generalization beyond training degradations, we first evaluate the model trained on the 3-degradation setting directly on the unseen desnowing task (\ie, zero-shot transfer) using the CSD dataset~\cite{chen2021all}. As shown in Tab.~\ref{tab:unseen}, \ourmethod extends effectively to this novel degradation without domain-specific tuning. 
We further conduct a real-world zero-shot evaluation on underwater images. \ourmethod-S achieves 16.78 dB PSNR and 0.770 SSIM, outperforming the best prior method MoCE-IR by +0.87 dB while being more compact (Tab.~\ref{table:underwaterIR}). Notably, our model has never seen underwater data during training, underscoring its robustness to unseen degradation scenarios.

\begin{table}[!t]
    \centering
    \footnotesize
    \setlength\tabcolsep{6pt}
    \setlength{\extrarowheight}{0.2pt}
    \caption{\textit{Complexity Analysis.} FLOPs are computed on image size $224\times224$ via a NVIDIA A100 (40G) GPU.}
    \label{tab:efficiency_overview}
    \begin{tabular}{lcccc}
        \toprule
        Method  & PSNR (dB, $\uparrow$) & Memory ($\downarrow$)  & Params. ($\downarrow$) & FLOPs ($\downarrow$) \\
        \midrule
        \rowcolor{gray!10}AirNet~\cite{li2022all} & 31.20 & 4829M &  8.93M  & 238G \\
        PromptIR~\cite{potlapalli2023promptir} & 32.06 & 9830M & 35.59M & 132G \\
        \rowcolor{gray!10}IDR~\cite{zhang2023ingredient} & - & 4905M & 15.34M & 98G \\
        MoCE-IR~\cite{zamfir2025moce} & 32.73 & 5887M & 25.35M & 80.59 \(\pm\) 5.21G \\
        \rowcolor{gray!10}MoCE-IR-S~\cite{zamfir2025moce} & 32.57 & \textcolor{tabred}{\textbf{4228M}} & 11.47M & 36.93 \(\pm\) 2.32G \\
        \midrule
        \ourmethod-T(\textit{ours}) & \textcolor{tabred}{\textbf{32.80}} & {{4969M}} & \textcolor{tabred}{\textbf{5.74M}} & \textcolor{tabred}{\textbf{26} \(\pm\) \textbf{1.98G}} \\ 
        \ourmethod-S(\textit{ours}) & \textcolor{tabred}{\textbf{32.86}} & {{6661M}} & \textcolor{tabred}{\textbf{8.51M}} & \textcolor{tabred}{\textbf{39} \(\pm\) \textbf{1.87G}} \\ 
        \bottomrule
    \end{tabular}
\end{table}

\noindent \textbf{Model efficiency.}
Tab.~\ref{tab:efficiency_overview} compares memory usage, FLOPs, and parameters across recent All-in-One IR methods. 
With our hybrid block design and the proposed degradation adaptation module, \ourmethod achieves a \textbf{0.74 dB} PSNR gain over the baseline PromptIR~\cite{potlapalli2023promptir}, while reducing 
parameters by \textbf{83.9\%}, and FLOPs to only \textbf{26G}, making it \textbf{80\%} more computationally efficient. 
Compared with MoCE-IR-S, \ourmethod lowers FLOPs by \textbf{29.73\%} (26G vs.~37G) and parameters by \textbf{49.96\%} (5.74M vs.~11.47M), while maintaining comparable or superior accuracy. 
Such reductions not only improve efficiency in terms of model size and computation, but also translate into a smaller energy and compute footprint during inference, which is particularly relevant for resource-constrained or edge deployment. This positions \ourmethod as a strong and sustainable baseline for future All-in-One IR research.

\begin{figure*}[t]
    \centering
    \includegraphics[width=0.99\linewidth]{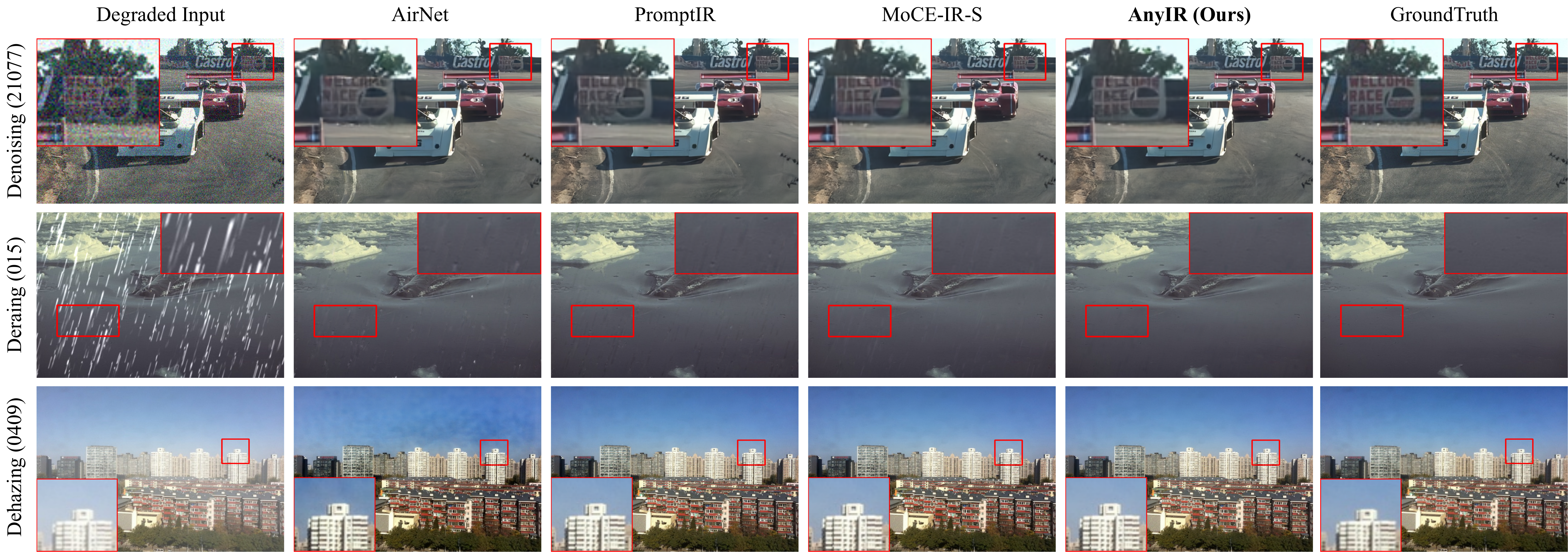}
    \caption{Visual comparison on three degradations. Zoom in for a better view.}
    \label{fig:exp:visual_results}
\end{figure*}

\begin{figure*}[t]
    \centering
    \includegraphics[width=0.97\linewidth]{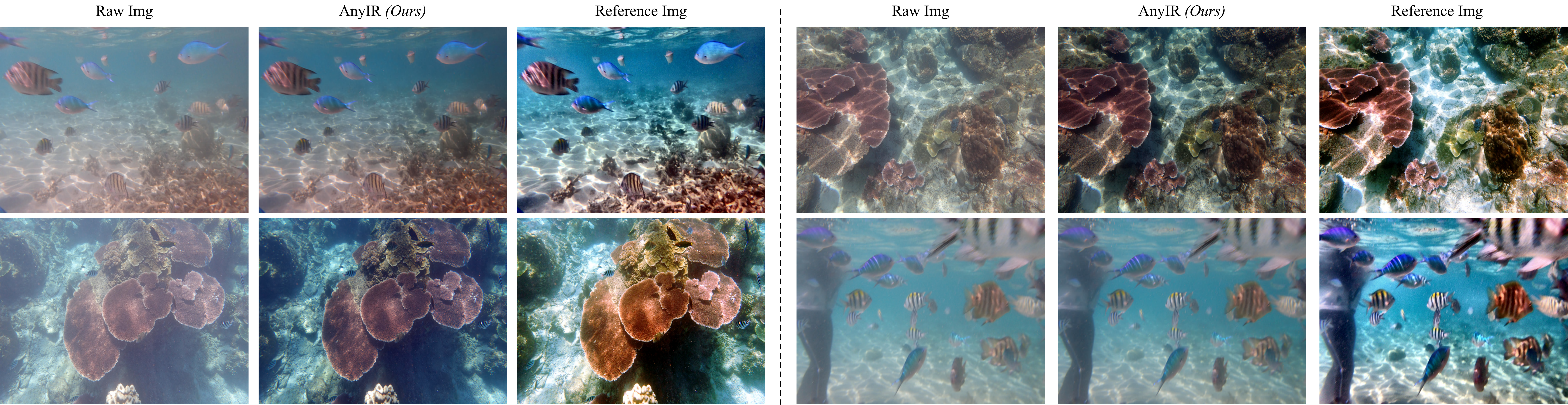}
    \caption{Zero-Shot Underwater Image Enhancement Visual Results. Zoom in for a better view.}
    \label{fig:exp:visual_results_zero}
\end{figure*}

\noindent \textbf{Visual results.}
To complement the quantitative results, we visualize the results of our method in Fig.~\ref{fig:exp:visual_results}. The visualizations demonstrate the efficacy of \ourmethod in dehazing, denoising, and draining, and we marked out the detailed region using the red boxes. 
In the dehazing task, AirNet~\cite{li2022all}, PromptIR~\cite{potlapalli2023promptir}, and MoCE-IR~\cite{zamfir2025moce} exhibit limitations in fully eliminating haziness, leading to noticeable color reconstruction discrepancies. In contrast, our \ourmethod effectively enhances visibility and ensures a precise color reconstruction. 
The denoising results also show that our \ourmethod can restore more detailed characters, demonstrating the rich texture edge recovery ability of our method. Meanwhile, in rainy scenes, previous methods continue to exhibit remnants of rain streaks. Please, \textbf{zoom in} for more details. In contrast, our approach excels at eliminating these artifacts and recovering underlying details, showing our superiority under adverse weather conditions.
Note that the PromptIR is approximately 6$\times$ larger than ours. Despite this, our method consistently produces visually superior results, demonstrating its effectiveness. 
Fig.~\ref{fig:exp:visual_results_zero} also shows that under the zero-shot setting, our method can also restore clear results.
More detailed visual examples are given in the appendix.

\begin{figure}[!t]
    \centering
    \includegraphics[width=0.7\linewidth]{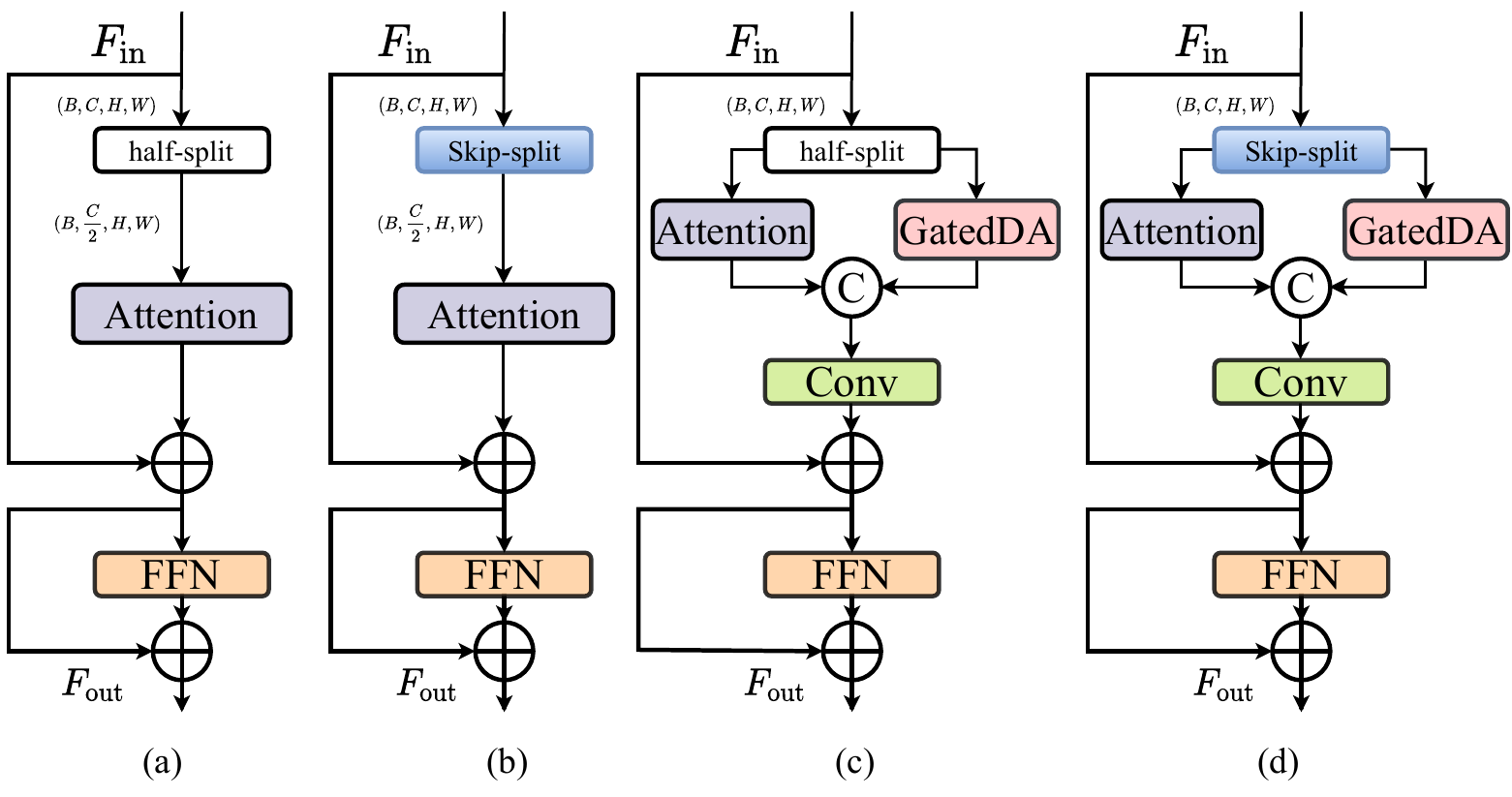}
    \caption{Structure of other different DAB variants during exploration.}
    \label{fig:ab_component}
\end{figure}

\subsection{Ablation Studies}
\label{ablation}
\noindent \textbf{Impact of different components.} We conduct detailed studies on the proposed components within the framework of \ourmethod. All experiments are conducted in the \textit{All-in-One} setting with \textit{three} degradations. We compare our simple feature-modeling block DAB against other variants.  As detailed in Tab.~\ref{tab:ablation}, we assess the effectiveness of our key architectural contributions by removing or replacing our designed module with its counterparts. The detailed architecture overview of these considered counterparts can be found in Fig.~\ref{fig:ab_component}. 

We first examine the impact of our skip-split operation (\textit{a}-\textit{b}), which yields a significant improvement of \textbf{1.02 dB} over the common half-split method. This validates and supports our motivation to deeply enable intertwining within the subparts of an input feature. 
Introducing our proposed GatedDA in parallel to the attention layer (\textit{c}) results in a substantial increase of $1.27$ dB. Combining GatedDA with the skip-split operation (\textit{d}) further enhances the reconstruction fidelity of our framework. This validates the effectiveness of our intention of using local gated details to reconstruct the degradation-aware output. Lastly, the introduction of cross-feature filtering (\textit{e}), please refer to Fig.~\ref{fig:framework} for our plain design, improves the interconnectivity between global-local features, further benefiting overall performance.
\begin{table}[!t]
    \centering
    \footnotesize
    \caption{Ablation comparison (Average PSNR) of the effect of each component under the 3-degradation.}
    \label{tab:ablation}
    \setlength{\extrarowheight}{0.5pt}
    \setlength\tabcolsep{6pt} 
    \begin{tabular}{lcccc}
        \toprule
        Method & Skip-Split & Fusion (Alg.~\ref{alg:sf_fusion}) & GatedDA & PSNR (dB, $\uparrow$) \\
        \midrule
        \rowcolor{gray!10} (a) & $\times$ & $\times$ & $\times$ & 30.85 \\
        (b) & \checkmark & $\times$ & $\times$ & 31.83 \\
        \rowcolor{gray!10} (c) & $\times$ & $\times$ & \checkmark & 32.13 \\
        (d) & \checkmark & $\times$ & \checkmark & 32.35 \\
        \rowcolor{gray!10} (e) & \checkmark & \checkmark & \checkmark & \textcolor{tabred}{\textbf{32.80}} \\
        \bottomrule        
    \end{tabular}
\end{table}

\begin{table}[!t]
    \centering
    \footnotesize
    \setlength\tabcolsep{10pt}
    \setlength{\extrarowheight}{1pt}
    \caption{Impact of different ($\alpha$, $\beta$, $\gamma$) settings on three-degradation.}
    \label{tab:ab_gatedDA}
    \begin{tabular}{lcc}
        \toprule
        ($\alpha$, $\beta$, $\gamma$) & PSNR $\uparrow$ & SSIM $\uparrow$ \\
        \midrule
        \rowcolor{gray!10}(0, \nicefrac{1}{2}, \nicefrac{1}{2}) & 32.14 & 0.910 \\
        (\nicefrac{1}{2}, 0, \nicefrac{1}{2}) & 32.21 & 0.912 \\
        \rowcolor{gray!10}(\nicefrac{1}{2}, \nicefrac{1}{2}, 0) & 31.37 & 0.907 \\
        (\nicefrac{1}{4}, \nicefrac{1}{4}, \nicefrac{1}{2}) & \textbf{\textcolor{tabred}{32.80}} & \textbf{\textcolor{tabred}{0.919}} \\
        \bottomrule
    \end{tabular}
\end{table}

\begin{figure}[!t]
    \centering
    \includegraphics[width=0.95\linewidth]{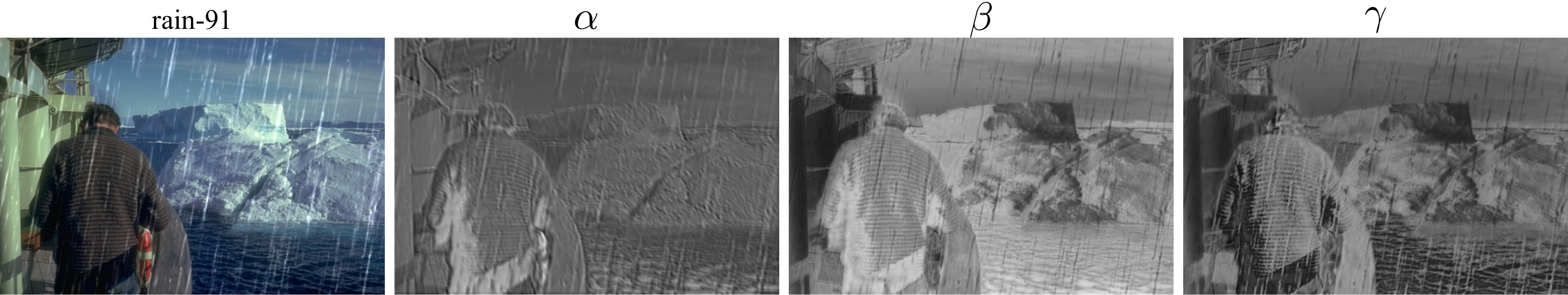}
    \caption{Visual feature maps of $\alpha$, $\beta$, and $\gamma$ within GatedDA.}
    \label{fig:ab_gic}
\end{figure}

\noindent \textbf{Impact of different $\alpha$, $\beta$, and $\gamma$ in gatedDA.} Tab.\ref{tab:ab_gatedDA} shows the different impact of different values of ($\alpha$, $\beta$, $\gamma$) in the gatedDA proposed (without Alg.~\ref{alg:sf_fusion}). We noticed that when setting one of these parameters to 0, the performance decreases. When we set ($\alpha$, $\beta$, $\gamma$) to ($\nicefrac{hidden}{4}$, $\nicefrac{hidden}{4}$, $\nicefrac{hidden}{2}$), the average performance increases. This means that each part of the proposed gatedDA is necessary. The visual features shown in Fig.~\ref{fig:ab_gic} indicate that $\alpha$, $\beta$, and $\gamma$ consistently focus on various aspects of the degraded regions, each specializing in different levels or types of degradation. See more analyses in our \textit{Supp. Mat.}

\begin{figure}[!t]
    \centering
    \includegraphics[width=0.95\linewidth]{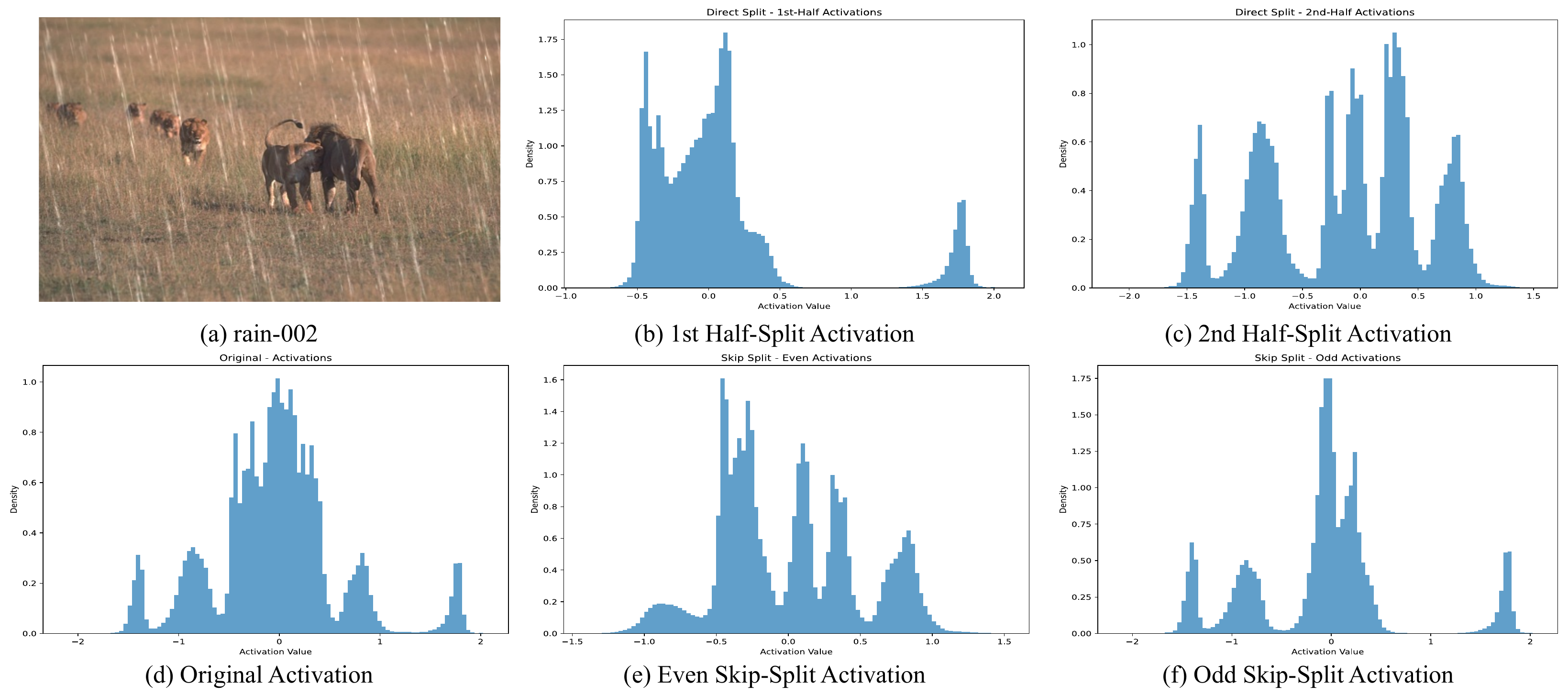}
    \caption{Visualization of the channel activation distribution.}
    \label{fig:vis_skip}
\end{figure}

\begin{table}[t]
    \centering
    \footnotesize
    \setlength\tabcolsep{4pt}
    \setlength{\extrarowheight}{0.5pt}
    \caption{\textit{Network Capacity and Complexity Analysis on \textbf{Deraining}.} Comparison of PSNR, model parameters, and FLOPs, highlighting the balance between network capacity and computational efficiency. FLOPs are computed for $224\times224$ input images using an NVIDIA Tesla A100 (40G) GPU.}
    \label{tab:supp_capacity}
    \begin{tabular}{lcccc}
        \toprule
        Method & Blocks & PSNR (dB, $\uparrow$) & Params. ($\downarrow$) & FLOPs ($\downarrow$) \\
        \midrule
        \rowcolor{gray!10} PromptIR~\cite{potlapalli2023promptir}(Base) & [4,6,6,8] & 37.04 & 35.6M & 132G \\
        Base (No Prompt) & [4,6,6,8] & 36.74 & 32.2M & 117G \\
        \rowcolor{gray!10} Base (Fix-Prompt) & [4,6,6,8] & 36.85 & - & - \\
        \midrule
        Base (No Prompt) & [3,5,5,7] & 36.84 & 29.1M & 101G \\
        \rowcolor{gray!10} Base (No Prompt) & [2,4,4,6] & 36.76 & 26.0M & 85G \\
        Base (No Prompt) & [1,3,3,5] & 36.63 & 22.9M & 68G \\
        \midrule
        \ourmethod(\textit{ours}) & [3,5,5,7] & \textcolor{tabred}{\textbf{37.99}} & \textcolor{tabred}{\textbf{5.74M}} & \textcolor{tabred}{\textbf{26G}} \\ 
        \bottomrule
    \end{tabular}
\end{table}

\noindent \textbf{Impact of the network capacity.} The results in Tab.~\ref{tab:supp_capacity}, derived from experiments on the draining task, provide valuable insight into the broader context of all-in-one IR. Although draining serves as a testbed, the findings reflect a key question in general-purpose restoration: \textit{Does increasing model capacity universally lead to better performance?} Our analysis suggests otherwise, emphasizing the importance of efficiency and task-aware design over brute-force scaling.

For example, the base PromptIR~\cite{potlapalli2023promptir} model with blocks [4,6,6,8] achieves a PSNR of 37.04 dB through 35.6M parameters and 132G FLOP, and progressively reducing capacity through configurations such as [3,5,5,7], [2,4,4,6] and [1,3,3,5] results in smaller models with lower computational costs but marginal decreases in PSNR. This highlights the diminishing returns of simply reducing complexity without optimization, as smaller models lose their capacity to capture the nuances of degradation factors.
In contrast, our \ourmethod validates that a careful balance between network capacity and architectural innovation can achieve SOTA results. With a similar block configuration ([3,5,5,7]) but a significantly optimized architecture, \ourmethod achieves a PSNR of 37.99 dB on the deraining task, outperforming larger models while using only 5.74M parameters and 26G FLOPs. This demonstrates the effectiveness of our task-specific improvements and highlights the potential to achieve better results with smaller, more efficient models.
These findings underscore a critical insight for all-in-one IR: 
\textit{While larger models may generalize better across diverse tasks, efficiency and tailored design can lead to both higher performance and practical utility}. 
The deraining results provide a compelling argument for rethinking model capacity in restoration frameworks, emphasizing that ``more'' is not always better, especially when thoughtful design can yield both performance gains and computational efficiency.

\begin{table}[!t]
    \centering
    \footnotesize
    \caption{\textit{Ablation comparison} of different fusion strategies.}
    \label{tab:ab_fusion}
    \setlength{\tabcolsep}{6pt}
    \setlength{\extrarowheight}{0.5pt}
    \begin{tabular}{lccc}
        \toprule
        Fusion & $\lambda$ & PSNR (dB, $\uparrow$) & SSIM ($\uparrow$) \\
        \midrule
        Spatial-only   & 1.0 & 32.37 & .915 \\
        Frequency-only & 0.0 & 32.09 & .919 \\
        Fixed (0.5)    & 0.5 & 32.63 & .917 \\
        Learnable      & --- & \textbf{32.80} & \textbf{.919} \\
        \bottomrule
    \end{tabular}
\end{table}
\noindent\textbf{Exploration of $\lambda$ value in Spatial-Frequency Fusion.} 
We explore different $\lambda$ settings in the fusion module: spatial-only, frequency-only, fixed, and learnable. As shown in Tab.~\ref{tab:ab_fusion}, the learnable strategy yields the best results, highlighting the benefit of adaptive spatial–frequency balancing. Moreover, both fixed and learnable schemes surpass single-branch counterparts, confirming the complementarity of the two domains.

\section{Discussion}
\label{sec:discuss}
\noindent\textbf{What does the proposed Skip-Split bring?}
Adjacent channels often contain redundant information due to spatial correlations in the data. Directly splitting the channels into two contiguous halves can lead to uneven feature distribution, with one half potentially capturing redundant features, while the other lacks important information. By using a simple skip-split method that interleaves channels between the two processing paths, we ensure a more balanced and diverse set of features in each path, enhancing the effectiveness of both the self-attention and convolutional components. These phenomena are also validated in the channel activation distribution visualization shown in Fig.~\ref{fig:vis_skip}.

\begin{figure}[!t]
    \centering
    \includegraphics[width=1.0\linewidth]{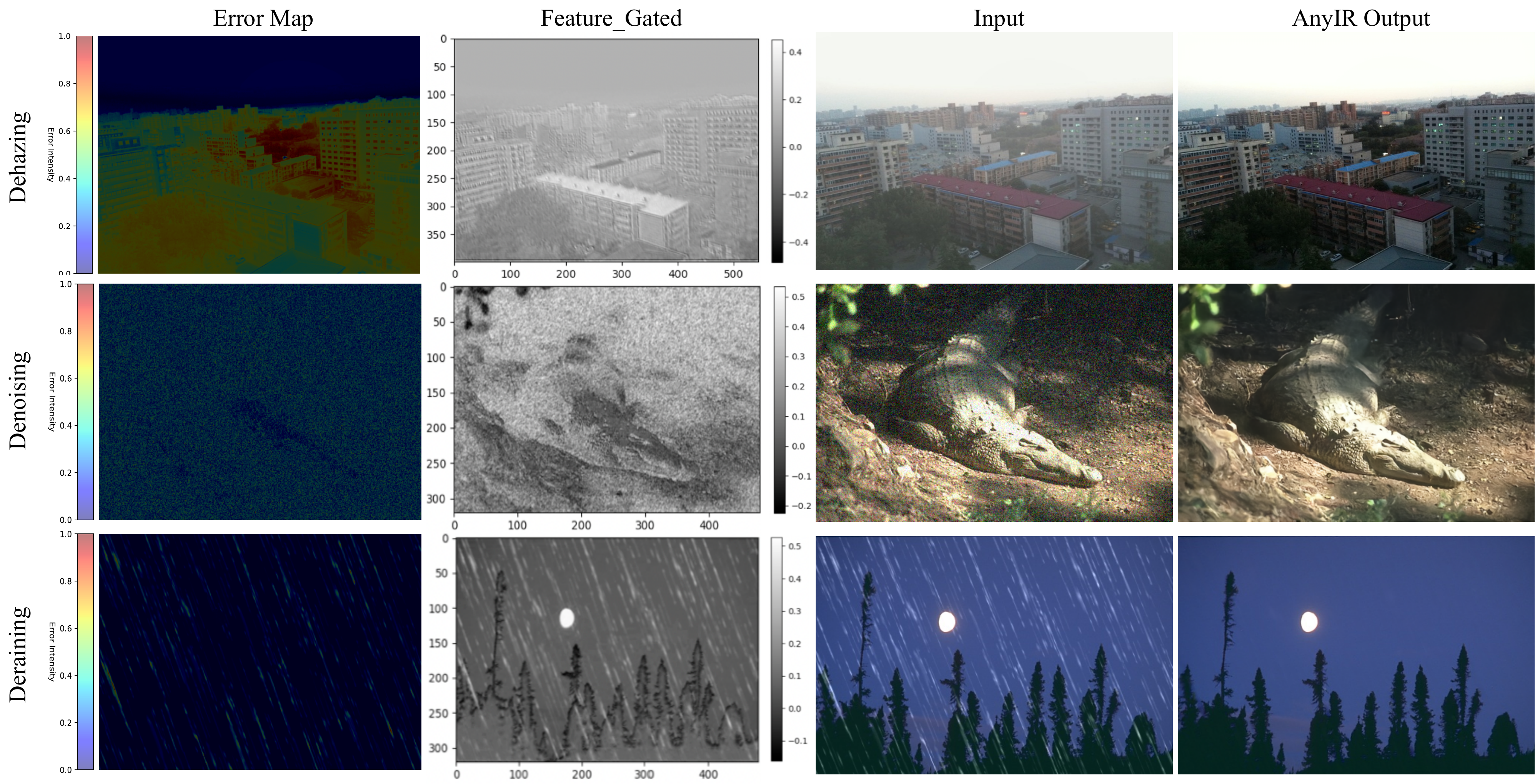}
    \caption{Error map and the output of GatedDA.}
    \label{fig:vis_gated}
\end{figure}

\noindent\textbf{Why does the proposed GatedDA \& the fusion Alg.~\ref{alg:sf_fusion} work?}
As shown in Fig.~\ref{fig:vis_gated}, the error map visualizations reveal that degradation in an input image is often unevenly distributed, manifesting itself as global patterns and localized clusters. This highlights the need for degradation modeling to capture both widespread and fine-grained distortions. The proposed GatedDA module addresses this by selectively activating in degraded regions, closely aligning with the actual degradation distribution, and effectively enhancing localized features. When combined with global attention, which captures broader contextual dependencies, the fusion algorithm (Alg.~\ref{alg:sf_fusion}) enables a more comprehensive understanding of the degradation structure of the image.

\begin{figure}[!t]
    \centering
    \includegraphics[width=1.0\linewidth]{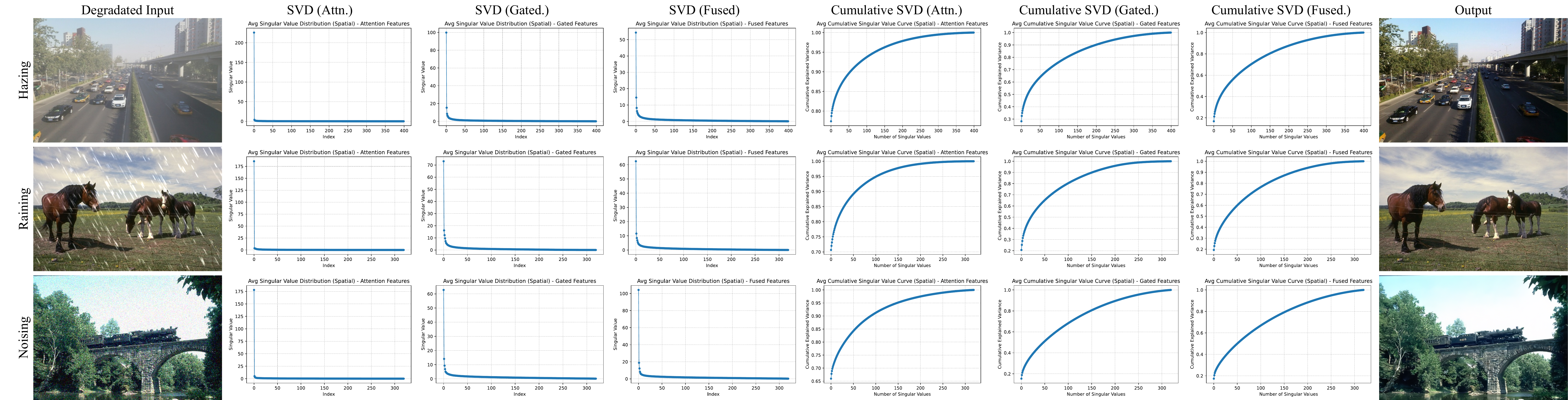}
    \caption{SVD and cumulative variance curves for attention, GatedDA, and the fused feature (Zoom in for a better view).}
    \label{fig:compare_demo}
\end{figure}

Further evidence is provided by the SVD and cumulative variance curves in Fig.~\ref{fig:compare_demo}, which demonstrate the complementary nature of the two modules. While the attention branch captures dominant global variations reflected in a steep variance accumulation and concentrated singular values, GatedDA captures more diverse and spatially distributed local signals. The fused representation achieves a better balance, integrating both local and global characteristics with faster variance accumulation than GatedDA alone. These results validate that the synergy between GatedDA and attention not only improves restoration quality but also enhances robustness across diverse types of degradation.
Although some aspects of this analysis can be interpreted at a more abstract representation-learning level, our formulation and conclusions are confined to the image restoration setting, where the model is developed and evaluated specifically for multi-degradation IR.

\noindent\textbf{Why is \ourmethod efficient?} \ourmethod reduces computational costs by splitting input channels: one half is processed by self-attention, the other by a Gated CNN block. This division reduces the complexity of self-attention from \(O(B \cdot \text{head} \cdot (H \cdot W)^2)\), while Gated CNN processes the remaining channels with a lower complexity of \(O(B \cdot C \cdot H \cdot W)\), especially beneficial for high-resolution inputs where \((H \cdot W)^2\) dominates. Furthermore, \ourmethod employs dimensionality reduction and parameter-efficient designs, collectively reducing the GFLOPs and model parameters. This balance of global context modeling and efficient local feature extraction enables \ourmethod to minimize computational costs without sacrificing performance.

\noindent\textbf{Is scaling down the new advantage?} Although recent methods in image restoration often scale up model size and complexity, our \ourmethod takes a different path: scaling down. Instead of relying on large architectures, we embrace a simple but non-trivial design, using targeted components like the GatedDA module to capture degradation without excessive parameters effectively. This simplicity is not a compromise; it is an asset - yielding high-quality restoration with minimal computational demand, faster training, and greater adaptability.
The All-in-One IR framework, though new compared to degradation-specific methods, faces a key limitation: an imbalance in data distribution, with certain degradations (\eg dehazing) dominating. Our experiments suggest that a more balanced dataset could significantly improve performance across degradation types, offering a constructive direction for future work. 
We note that these discussions and observations are made strictly within the scope of multi-degradation image restoration, and are not intended to imply a general-purpose image processing or prediction framework beyond the IR domain.

\noindent\textbf{Effect of degradation distribution.}
We further observe that commonly used multi-degradation IR training sets (3 Degradation) tend to exhibit an imbalanced composition across degradation types. 
By making the distribution more uniform (Fig.~\ref{fig:samples}) under the same 3-degradation setting—without increasing the total number of samples—we consistently obtain improved restoration performance (Tab.~\ref{tab:ab_data}). 
This trend suggests that the learned reconstruction prior is influenced by the training degradation distribution, and a more balanced composition can lead to more stable and generalized restoration behavior.

\section{Conclusion}
\label{sec:conclusion}
We introduced AnyIR, an efficient multi-degradation image restoration model that unifies the degradations considered in current All-in-One IR settings within a single framework.
By combining gated guidance with a spatial–frequency fusion strategy, the model learns embeddings that capture both degradation-specific cues and invariant structures, enabling robust restoration across tasks. 
Extensive experiments, including evaluations on unseen and real-world degradations, show that \ourmethod achieves state-of-the-art accuracy with substantially reduced parameters and FLOPs, while maintaining strong generalization ability. 
We believe that \ourmethod provides a strong and efficient baseline for future research, advancing both the practical deployment and the broader understanding of learning-based all-in-one image restoration.

\subsubsection*{Acknowledgments} This work was partially supported by the FIS project GUIDANCE (Debugging Computer Vision Models via Controlled Cross-modal Generation) (No. FIS2023-03251), the Alexander von Humboldt Foundation, and the National Natural Science Foundation of China (62403345).

\bibliography{main}
\bibliographystyle{tmlr}

\appendix
\section{Experimental Protocols}
\label{suppsec:exp_setup}
\subsection{Datasets}
\textbf{3 Degradation Datasets.}
For both the All-in-One and single-task settings, we follow the evaluation protocols established in prior works~\cite{li2022all,potlapalli2023promptir,zamfir2025moce}, utilizing the following datasets:
For image denoising in the single-task setting, we combine the BSD400~\cite{arbelaez2010contour} and WED~\cite{ma2016waterloo} datasets, and corrupt the images with Gaussian noise at levels $\sigma \in \{15, 25, 50\}$. 
BSD400 contains 400 training images, while WED includes 4,744 images. 
We evaluate the denoising performance on BSD68~\cite{martin2001database} and Urban100~\cite{huang2015single}.
For single-task deraining, we use Rain100L~\cite{yang2020learning}, which provides 200 clean/rainy image pairs for training and 100 pairs for testing.
For single-task dehazing, we adopt the SOTS dataset~\cite{li2018benchmarking}, consisting of 72,135 training images and 500 testing images.
Under the All-in-One setting, we train a unified model on the combined set of the aforementioned training datasets for 120 epochs and directly test it across all three restoration tasks.

\textbf{5 Degradation Datasets.}
The 5-degradation setting is built upon the 3-degradation setting, with two additional tasks included: deblurring and low-light enhancement. 
For deblurring, we adopt the GoPro dataset~\cite{nah2017deep}, which contains 2,103 training images and 1,111 testing images. 
For low-light enhancement, we use the LOL-v1 dataset~\cite{wei2018deep}, consisting of 485 training images and 15 testing images. 
Note that for the denoising task under the 5-degradation setting, we report results using Gaussian noise with $\sigma = 25$. The training takes 130 epochs.

\textbf{Composited Degradation Datasets.}
Regarding the composite degradation setting, we use the CDD11 dataset~\cite{guo2024onerestore}. CDD11 consists of 1,183 training images for:
\textit{(i) 4 kinds of single-degradation types:} haze (H), low-light (L), rain (R), and snow (S);
\textit{(ii) 5 kinds of double-degradation types:} low-light + haze (l+h), low-light+rain (L+R), low-light + snow (L+S), haze + rain (H+R), and haze + snow (H+S).
\textit{(iii) 2 kinds of Triple-degradation type:} low-light + haze + rain (L+H+R), and low-light + haze + snow (L+H+S).
We train our method for 150 epochs (significantly fewer than the 200 epochs used in MoCE-IR~\cite{zamfir2025moce}), and we keep all other settings unchanged.

\textbf{Zero-Shot Underwater Image Enhancement Dataset.}
For the zero-shot underwater image enhancement setting, we follow the evaluation protocol of DCPT~\cite{hu2025universal} by directly applying our model, trained under the 5-degradation setting, on the UIEB dataset~\cite{li2019underwater} without any finetuning. 
UIEB consists of two subsets: 890 raw underwater images with corresponding high-quality reference images, and 60 challenging underwater images. 
We evaluate our zero-shot performance on the 890-image subset with available reference images.

\subsection{Implementation Details}
\label{suppsec:implementation}
Our \ourmethod framework is designed to be end-to-end trainable, eliminating the need for multi-stage optimization of individual components. The architecture features a robust 4-level encoder-decoder structure, characterized by varying numbers of Degradation Adaptation Blocks (DAB) at each level, specifically $[3, 5, 5, 7]$ from highest to lowest level. Following established practices \cite{potlapalli2023promptir,zamfir2025moce}, we conducted training over $130$ epochs via a batch size of $32$ for the All-in-One and mixed settings. Optimization employed the $L_{1}$ loss and Fourier with the Adam optimizer \cite{kingma2015adam} (initial learning rate of $\num{2e-4}$, $\beta_1=0.9$, $\beta_2=0.999$) and cosine decay schedule. During training, we employed random crops of size $128^2$ and applied horizontal and vertical flips as augmentations. All experiments were performed using 2 NVIDIA Tesla A100 (40G) GPUs.

We also propose two scaled variants of our \ourmethod, namely Tiny (\ourmethod-T) and Small (\ourmethod-S). As detailed in Tab.~\ref{tab:supp:model_details}, these variants differ in terms of the number of Degradation Adaption Blocks (DAB) across scales, the input embedding dimension, the FFN expansion factor, and the number of refinement blocks.
\begin{table}[!t]
    \centering
    \scriptsize
    \vspace{-2mm}
    \caption{The details of the tiny and small versions of the proposed \ourmethod.}
    \setlength{\extrarowheight}{0.1pt}
    \setlength\tabcolsep{16pt} 
    \label{tab:supp:model_details}
    \scalebox{1.0}{
    \begin{tabular}{l|c|c}
    \toprule
     & \ourmethod-Tiny & \ourmethod-Small  \\ 
     \midrule
    The Number of the DAB crosses 4 scales & [3, 5, 5, 7] & [4, 6, 6, 8]  \\
    The Input Embedding Dimension& 28 & 32 \\
    The FFN Expansion Factor & 2 & 2 \\
    The Number of the Refinement Blocks & 4 & 4 \\
    \midrule
    Params. ($\downarrow$) & 5.74M  & 8.51M  \\
    FLOPs ($\downarrow$) & 26G & 39 G  \\
    \bottomrule
    \end{tabular}
    }
\end{table}

\section{Discussion And Analysis}
\label{suppsec:discuss}
\noindent \textbf{The visual illustration of Skip-Split and what it brings.} 
Fig.~\ref{fig:skip_pca} presents a 3D PCA visualization comparing Half-Split and our proposed Skip-Split. The results show that Skip-Split yields a more uniform and well-spread feature distribution, suggesting stronger and more discriminative representations. For a more intuitive understanding, a side-by-side visual comparison is also provided in the right part of Fig.~\ref{fig:skip_pca}.

\begin{figure*}[tp]
    \centering
    \includegraphics[width=1\linewidth]{figs/skip_pca.pdf}
    \caption{3D PCA Visualization between the Half-Split and our proposed Skip-Split, and the visual illustration of Skip-Split.}
    \label{fig:skip_pca}
\end{figure*}

\begin{figure}[tp]
    \centering
    \includegraphics[width=0.99\linewidth]{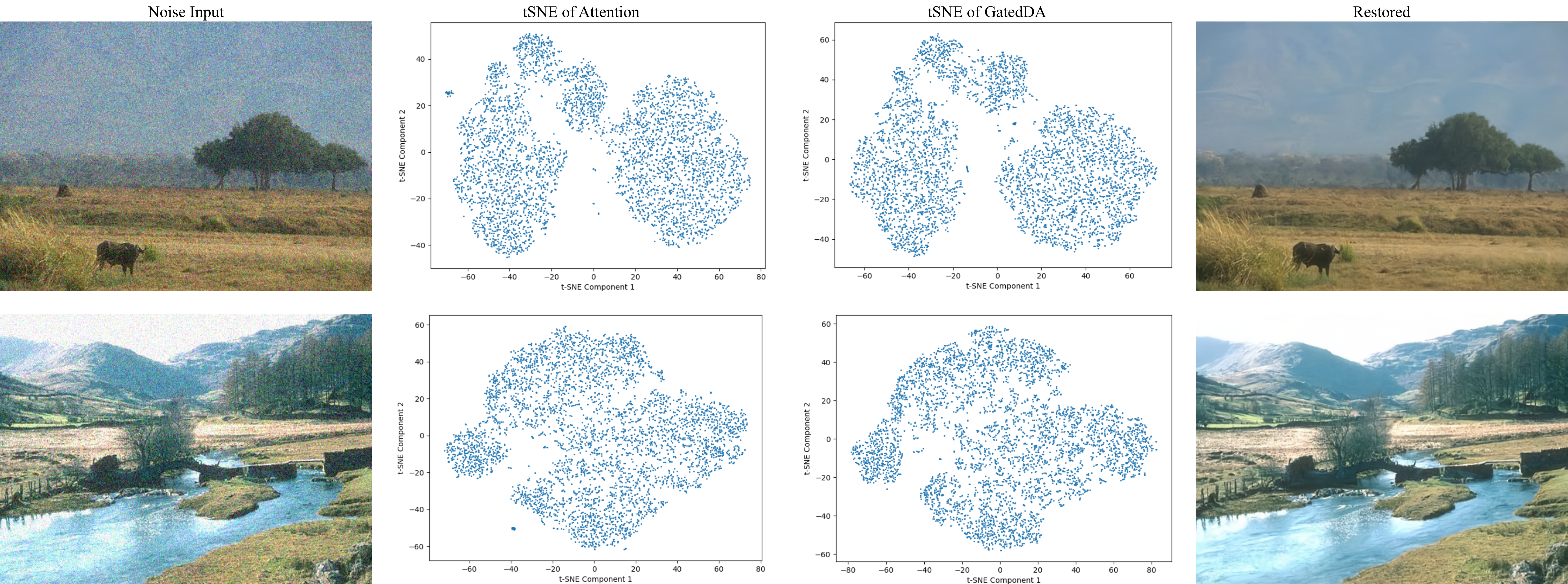}
    \caption{tSNE comparison between attention and GatedDA of the proposed method.}
    \label{fig:supp_tsne}
\end{figure}

\noindent \textbf{What does GatedDA bring that differs from attention?}
As the visualization result shown in Fig.~\ref{fig:supp_tsne}, the t-SNE visualizations of the feature maps reveal a key distinction between GatedDA and attention mechanisms. While attention effectively captures global relationships and context, its feature representations tend to exhibit less structural separation in the embedding space. In contrast, GatedDA introduces a localized focus on degradation-specific regions, resulting in a more distinct clustering of features in the t-SNE space. This separation highlights GatedDA's ability to emphasize degradation-specific information, complementing the broader scope of attention. Together, this synergy allows GatedDA to enhance image restoration by targeting specific degradation patterns while leveraging the global context provided by attention. As evidenced in the restored outputs, this combination leads to a more accurate recovery of both global and local details.

\begin{figure*}[tp]
    \centering
    \includegraphics[width=0.97\linewidth]{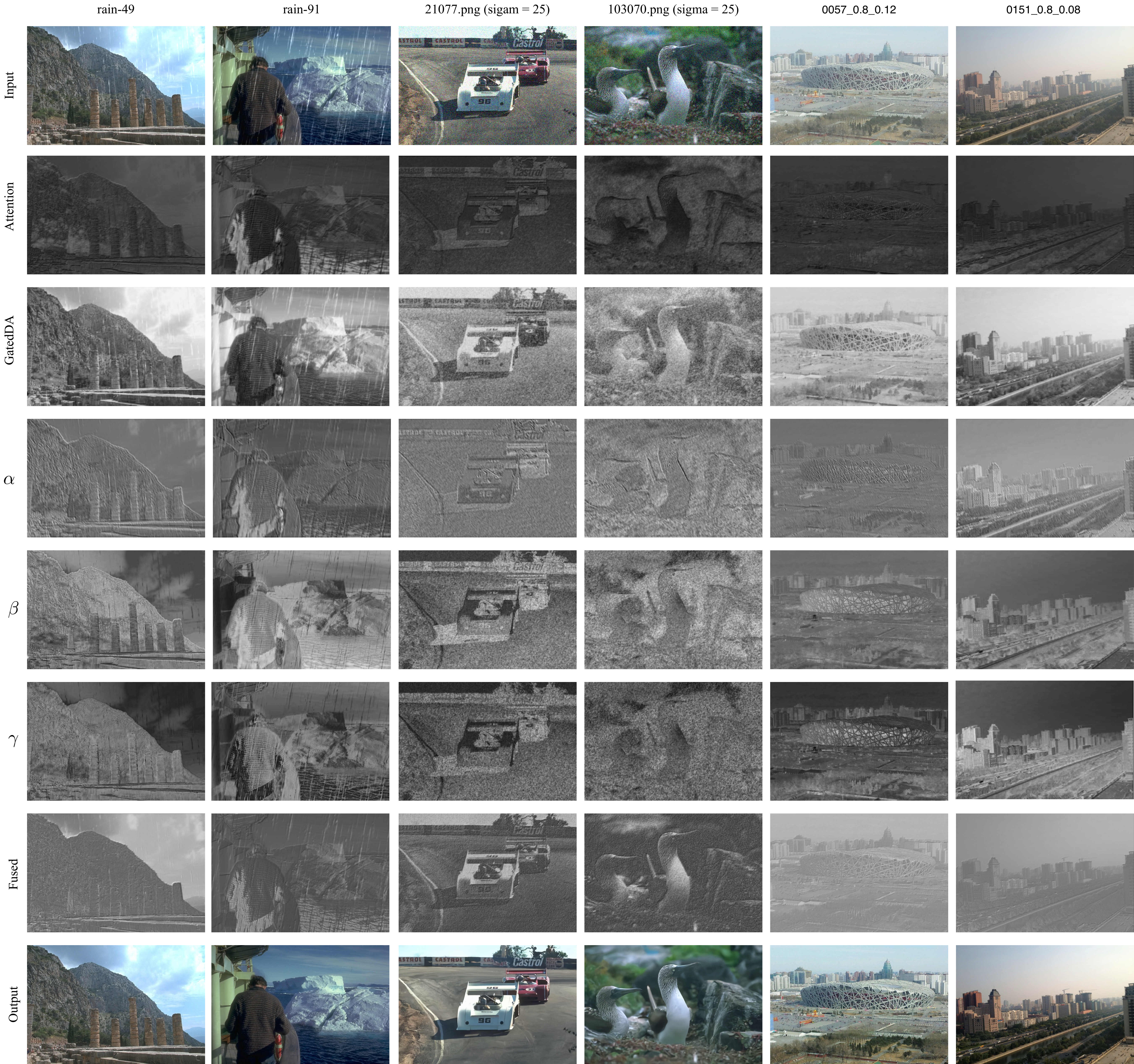}
    \caption{Visualization of the feature maps of each part within the proposed Degradation Adaptation Block (DAB).}
    \label{fig:feats_vis}
\end{figure*}

\noindent \textbf{Attention and GatedDA Feature Maps.}  
To investigate the effectiveness of the proposed Degradation Adaptation Block (DAB), we provide a detailed visualization of key components in Fig.~\ref{fig:feats_vis}. These include the attention mechanism, the GatedDA module, and its components—$\alpha$, $\beta$, and $\gamma$—along with the fused feature map (combining attention and GatedDA), and the final restored image.

From the visualizations of the attention feature map in the second row of Fig.~\ref{fig:feats_vis} and the GatedDA feature map in the third row, it is evident that GatedDA effectively models degradation factors such as rain, noise, and haze. This demonstrates the capability of GatedDA to capture and emphasize degradation-related information automatically.

\noindent \textbf{Role of $\alpha$, $\beta$, and $\gamma$ in Degradation Modeling.}
To further analyze how $\alpha$, $\beta$, and $\gamma$ within GatedDA contribute to degradation modeling, we refer to the visualizations in the 4th to 6th rows of Fig.~\ref{fig:feats_vis}. These maps show that $\alpha$, $\beta$, and $\gamma$ consistently focus on various aspects of the degraded regions, each specializing in different levels or types of degradation. This specialization allows the model to adapt to varying degrees and types of degradation, making it more versatile and effective in the context of Image Restoration (IR) under a "One-for-Any" setting.

\noindent \textbf{Fused Attention and GatedDA Features.} 
The fused feature map, combining attention and GatedDA, offers further insights (visualized in Fig.~\ref{fig:feats_vis}). Compared to the original attention map, the fused map exhibits a richer representation of degradation, which equips the model with more comprehensive information for accurate restoration.
In conclusion, the proposed DAB enhances the attention mechanism by embedding richer degradation information, thus improving restoration quality. The GatedDA module within DAB introduces flexibility and diversity in handling various types and levels of degradation, contributing to the overall robustness and effectiveness of our method.

\begin{figure}[tp]
    \centering
    \includegraphics[width=0.8\linewidth]{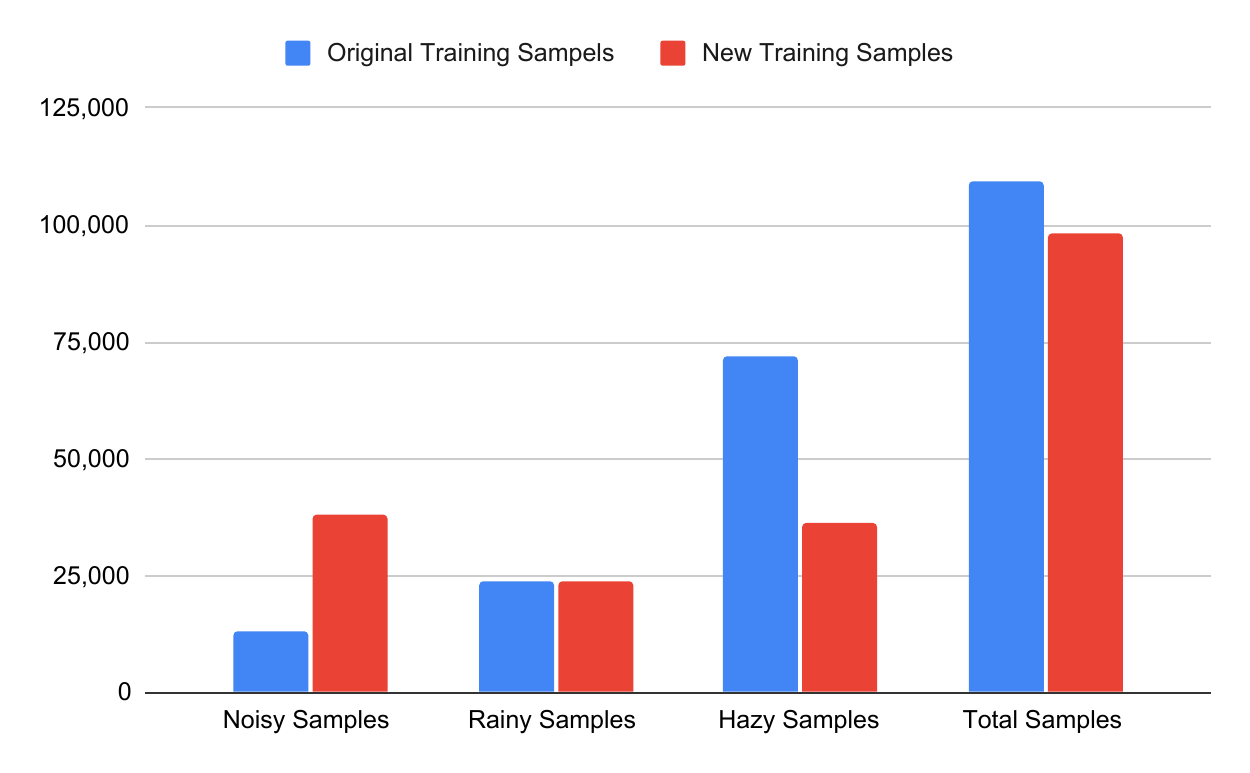}
    \caption{Training samples statistics.}
    \label{fig:samples}
\end{figure}

\noindent \textbf{Affect of data distribution.}
As illustrated in Fig.~\ref{fig:samples}, the original training samples exhibit significant variation across different IR tasks. Based on this observation, we propose a new set of training samples that over-represent the denoising samples and reduce the dehazing samples by half. 
This adjustment aims to achieve a more balanced data distribution. 
The results, presented in Tab.~\ref{tab:ab_data}, demonstrate that training \ourmethod with this revised set of samples, even when using only 90\% of the original total training samples, yields significant performance improvements.

\begin{table}[!t]
    \centering
    \footnotesize
    \caption{\textbf{Average} results for \textit{denoising} and \textit{overall} (\ie dehazing, deraining, and denoising) under the 3-degradation IR setting.}
    \label{tab:ab_data}
    \vspace{-1mm}
    \setlength{\tabcolsep}{6pt}
    \setlength{\extrarowheight}{0.5pt}
    \begin{tabular}{lcccc}
        \toprule
        Method & \multicolumn{2}{c}{Denoising} & \multicolumn{2}{c}{Overall} \\ 
        \midrule
        PromptIR~\cite{potlapalli2023promptir} & 31.11 & \cellcolor{clblue!50}.873 & 32.06 & \cellcolor{clblue!50}.913 \\
        \midrule
        \ourmethod (Original, ours) & 31.26 & \cellcolor{clblue!50}.878 & 32.80 & \cellcolor{clblue!50}.920 \\
        \ourmethod (New, ours) & \textbf{31.31} & \cellcolor{clblue!50}\textbf{.879} & \textbf{32.89} & \cellcolor{clblue!50}\textbf{.921} \\  
        \bottomrule        
    \end{tabular}
\end{table}

\begin{table}[!t]
    \centering
    \footnotesize
    \setlength{\tabcolsep}{6pt}
    \setlength{\extrarowheight}{2pt}
    \caption{Full \textbf{inference time} comparison under the 3-degradation IR setting. The lower the better.}
    \vspace{-2mm}
    \label{tab:inferencetime}
    \begin{tabular}{lccc}
        \toprule
        Task & Num. Samples & PromptIR~\cite{potlapalli2023promptir} & \ourmethod (Ours) \\
        \midrule
        Denoising ($\sigma=15$) & 68 & 41s & 30s \\ 
        Denoising ($\sigma=25$) & 68 & 41s & 30s \\ 
        Denoising ($\sigma=50$) & 68 & 42s & 30s \\ 
        \midrule
        Deraining & 100 & 60s & 43s \\ 
        \midrule
        Dehazing & 500 & 416s & 274s \\
        \midrule
        Frames Per Second & - & 1.34 & 1.97 \\
        \bottomrule        
    \end{tabular}
    \vspace{0mm}
\end{table}

\section{Additional Results}
\label{suppsec:add_exp}
\noindent\textbf{Full Inference Time Comparison.}
Table~\ref{tab:inferencetime} provides a detailed comparison of the full inference time between our method, \ourmethod, and PromptIR~\cite{potlapalli2023promptir} under the 3-degradation all-in-one IR setting. Across all tasks, our method demonstrates significantly faster inference, highlighting its computational efficiency. For denoising tasks with varying noise levels ($\sigma=15$, $25$, $50$), \ourmethod reduces the inference time from 41-42 seconds to just 30 seconds, achieving over a 25\% improvement. Similarly, for deraining, \ourmethod processes 100 samples in 43 seconds, compared to 60 seconds for PromptIR, and for dehazing, it processes 500 samples in 274 seconds, substantially faster than PromptIR's 416 seconds. 
In terms of frames per second (FPS), \ourmethod achieves 1.97 FPS, outperforming PromptIR’s 1.34 FPS by nearly 50\%. These results emphasize the efficiency of \ourmethod in handling various degradation tasks, making it highly suitable for practical applications where computational speed is critical without compromising performance.

These results indicate that our \ourmethod model not only maintains high performance but also offers substantial efficiency improvements, making it more suitable for real-time applications on resource-constrained devices. The reduced inference time ensures faster processing, enabling seamless deployment in scenarios requiring rapid decision-making, such as real-time video restoration or on-device image enhancement. Additionally, the improvement in frames per second (FPS) demonstrates its practicality for large-scale datasets and streaming applications. By achieving an effective balance between accuracy and speed, \ourmethod provides a compelling solution for efficient, high-performance image restoration.

\section{Additional visual results.}
\label{suppsec:vis}
\begin{figure}[tp]
    \centering
    \includegraphics[width=0.99\linewidth]{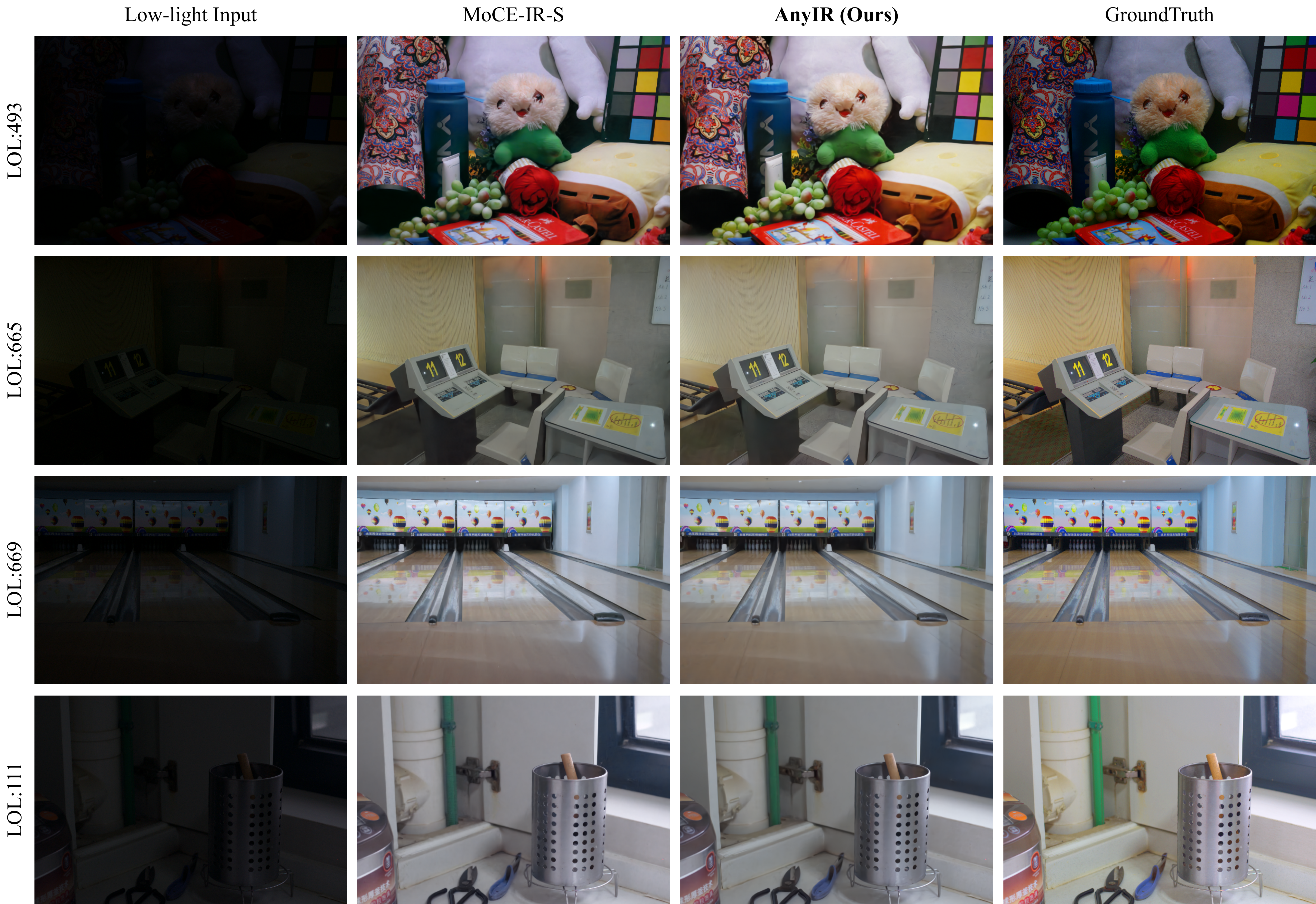}
    \caption{Additional visual comparison for low-light enhancement under the 5-degradation setting.}
    \label{fig:supp_visual_ll}
\end{figure}

\begin{figure*}[tp]
    \centering
    \includegraphics[width=0.99\linewidth]{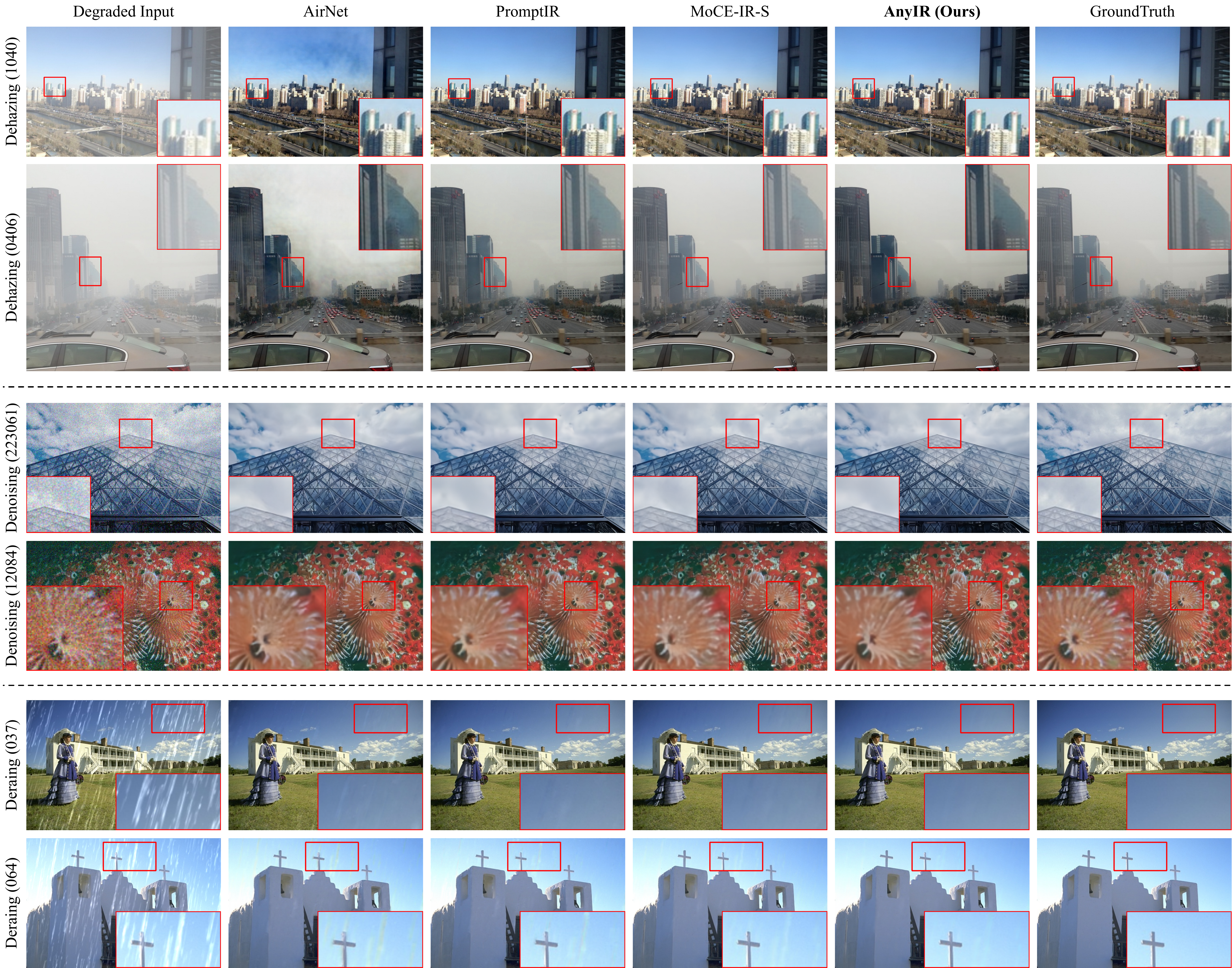}
    \caption{Additional visual comparison under the 3-degradation setting.}
    \label{fig:supp_visual}
\end{figure*}

The low-light enhanced visual comparison is provided in Fig.~\ref{fig:supp_visual_ll}.
The visual comparison results under the 3-degradation IR setting are shown in Fig.~\ref{fig:supp_visual}. 
It shows that the proposed \ourmethod can effectively restore the clean image from its degraded counterparts compared to other comparison methods.

\section{Limitations and Future Work}
\label{suppsec:limit}

Although the proposed \ourmethod achieves strong performance and efficiency under the current all-in-one image restoration (IR) setting, it also presents several limitations that define the scope of its applicability.

A first limitation lies in the imbalance of degradation distribution in the benchmark training data. Certain degradations (\eg, haze or rain) appear more frequently or with wider variation than others, which may bias the model toward better performance on dominant categories while providing smaller gains on under-represented ones. In future work, we plan to investigate more balanced or curriculum-style degradation scheduling, as well as data reweighting strategies, in order to improve robustness across heterogeneous degradation regimes.

In addition, while GatedDA and the spatial–frequency fusion mechanism provide consistent improvements across mixed and spatially non-uniform degradations, they are not universally optimal under all conditions. Since GatedDA selectively amplifies locally degraded regions, its benefit is weaker when degradations are globally uniform and largely dominated by global corruption. Likewise, our fusion design assumes complementary cues across spatial and frequency domains; when the degradation is strongly biased toward a single domain (\eg, purely frequency-structured noise), the contribution of the other branch may be limited. Furthermore, the efficiency of our design arises from reducing attention channel capacity and compensating it with lightweight local modeling, which trades some global modeling flexibility for improved computational economy. Exploring adaptive routing between attention and GatedDA, as well as task-dependent fusion weighting, is a promising direction to mitigate these trade-offs.

Finally, we clarify that in this work, ``All-in-One'' refers to a single model jointly trained across the degradation types covered in our benchmark setting, rather than an unrestricted universal solution for arbitrary degradations beyond this scope.

\section{Broader Impact}
\label{suppsec:impact}
The development of our unified image restoration model has significant potential, extending its impact beyond technical advancements. By reducing model complexity and computational requirements, our approach makes high-quality image restoration accessible on resource-constrained platforms like mobile devices. This enables efficient restoration in fields such as telemedicine, remote sensing, and digital archiving. Additionally, minimizing the computational footprint reduces the environmental impact of large-scale data processing, aligning with sustainable computing practices. The public availability of our code will further foster innovation and collaboration within the scientific community, setting new standards for efficiency and expanding practical applications in image restoration.

\end{document}